\documentclass{article} 
\usepackage{iclr2027_conference,times}

\usepackage{amsmath,amsfonts,bm}

\def\eqref#1{equation~\ref{#1}}

\def\1{\bm{1}}

\DeclareMathAlphabet{\mathsfit}{\encodingdefault}{\sfdefault}{m}{sl}
\SetMathAlphabet{\mathsfit}{bold}{\encodingdefault}{\sfdefault}{bx}{n}

\newcommand{\E}{\mathbb{E}}

\newcommand{\R}{\mathbb{R}}

\DeclareMathOperator*{\argmax}{arg\,max}

\usepackage[utf8]{inputenc} 
\usepackage[T1]{fontenc}    
\usepackage{booktabs}       
\usepackage{amsfonts}       
\usepackage{nicefrac}       
\usepackage{microtype}      
\usepackage{xcolor}         
\usepackage{microtype}
\usepackage{graphicx}
\usepackage{tikz}
\usepackage{caption}
\usepackage[toc,page,header]{appendix}
\usepackage{titletoc}
\usepackage{amsmath}
\usepackage{amssymb}
\usepackage{mathtools}
\usepackage{amsthm}
\theoremstyle{plain}\newtheorem{proposition}{Proposition}\newtheorem{lemma}{Lemma}\newtheorem{corollary}{Corollary}
\theoremstyle{remark}\newtheorem{remark}{Remark}\theoremstyle{plain}
\usepackage[table]{xcolor}
\usepackage{adjustbox}
\usepackage{pdfpages}
\usepackage{pgfplots}
\usepackage{listings}
\usepackage{algorithm}
\usepackage{algorithmic}
\usepackage{wrapfig}
\usepackage{enumitem}
\usepackage{tabularew}
\usepackage{subcaption}
\usepackage{minitoc}
\usepackage[most]{tcolorbox}
\usepackage{comment}
\usepackage[toc,page,header]{appendix}
\usepackage{booktabs} 

\newcommand{\ours}{\textit{MSPR}} 
\newcommand{\vamn}{\texttt{v-antmaze-medium-navigate}}
\newcommand{\vams}{\texttt{v-antmaze-medium-stitch}}

\newcommand{\valn}{\texttt{v-antmaze-large-navigate}}
\newcommand{\vals}{\texttt{v-antmaze-large-stitch}}

\newcommand{\vcsp}{\texttt{v-cube-single-play}}

\newcommand{\vsp}{\texttt{v-scene-play}}
\newcommand{\vptp}{\texttt{v-puzzle-3x3-play}}

\definecolor{ourcolor}{HTML}{7377FA}

\pgfplotsset{compat=1.18}
\definecolor{sb_gray}{RGB}{127,127,127}

\definecolor{codegreen}{rgb}{0,0.6,0}
\definecolor{codegray}{rgb}{0.5,0.5,0.5}
\definecolor{codepurple}{rgb}{0.58,0,0.82}
\definecolor{backcolour}{rgb}{0.95,0.95,0.92}
\definecolor{sb_blue}{RGB}{31,119,180}
\definecolor{sb_orange}{RGB}{255,127,14}
\definecolor{sb_green}{RGB}{44,160,44}
\definecolor{sb_red}{RGB}{214,39,40}
\definecolor{sb_purple}{RGB}{148,103,189}
\definecolor{sb_brown}{RGB}{140,86,75}
\definecolor{sb_pink}{RGB}{227,119,194}
\definecolor{sb_gray}{RGB}{127,127,127}
\definecolor{sb_yellow}{HTML}{bcbd22}
\definecolor{sb_cyan}{RGB}{23,190,207}

\definecolor{mcgillred}{RGB}{237, 27, 47}

\lstdefinestyle{mystyle}{
    language=Python,
    xleftmargin=5.0ex,
    basicstyle=\footnotesize\ttfamily\linespread{4},
    backgroundcolor=\color{gray!10},
    commentstyle=\color{gray},
    alsoletter={<>-0123456789},
    morekeywords={self, as, from, 512, 256, 32, 3, 2, 1, 255, 0, 5, -1, 2, 1568, 10},
    ndkeywords={nn, F, torch, partial},
    ndkeywordstyle=\color{sb_orange},
    keywordstyle=\color{sb_blue},
    emph={import, def, return, if, else},
    emphstyle=\bfseries\color{sb_blue},
    numberstyle=\footnotesize\ttfamily\color{gray},
    stringstyle=\color{sb_blue},
    breakatwhitespace=false,
    breaklines=true,
    keepspaces=true,
    numbers=left,
    numbersep=5pt,
    showspaces=false,
    showstringspaces=false,
    showtabs=false,
    tabsize=2
}
\newcommand{\cutsectionup}{\vspace{-2pt}}
\newcommand{\cutsectiondown}{\vspace{-3pt}}

\usepackage{hyperref}
\usepackage{url}

\title{MSPR: Multi-scale Predictive Representations for Goal-conditioned Reinforcement Learning}

\author{%
Valliappan Chidambaram Adaikkappan*\\
  \thanks{Correspondence to \texttt{valliappan.chidambaramadaikkappa@mail.mcgill.ca} . }
  Mila, McGill University
  \And
  Sai Rajeswar \\ 
  ServiceNow Research \\
  \And
  Pietro Mazzaglia \\ 
  Qualcomm Research \\
  \And
  ~~~~~~~~~~~~~~~~~~~~~~~~~~~~~~~~~~~~~~~~~~~~~~~~~~~~~~~~~~~~David Meger \\ 
  ~~~~~~~~~~~~~~~~~~~~~~~~~~~~~~~~~~~~~~~~~~~~~~~~~~~~~~~~~~~~Mila, McGill University\\
}

\iclrfinalcopy 
\begin{document}

\maketitle

\begin{abstract}
This paper investigates robust representation learning in offline goal-conditioned reinforcement learning (GCRL). Particularly in sparse reward scenarios, learning representations that align state and goal latents is a challenge, as the encoder can learn goal-agnostic features that destabilize policy learning. We address this issue by learning the encoder's representation with alignment objectives that capture the environment across multiple scales, from local physical dynamics to long-horizon goal-directed structure. Concretely, we propose \ours{}, a framework that leverages multi-scale predictive supervision to enforce goal-directed alignment within the latent space. We demonstrate that \ours{} leads to strong performance on both vision and state-based tasks. Furthermore, we show that our approach is resilient under realistic, challenging data regimes, maintaining state-of-the-art performance across a wide variety of tasks.
\end{abstract}
\vspace{-3mm}
\begin{figure}[h]
    \centering
    \includegraphics[width=0.6\linewidth, trim=2mm 2mm 2mm 2mm, clip]{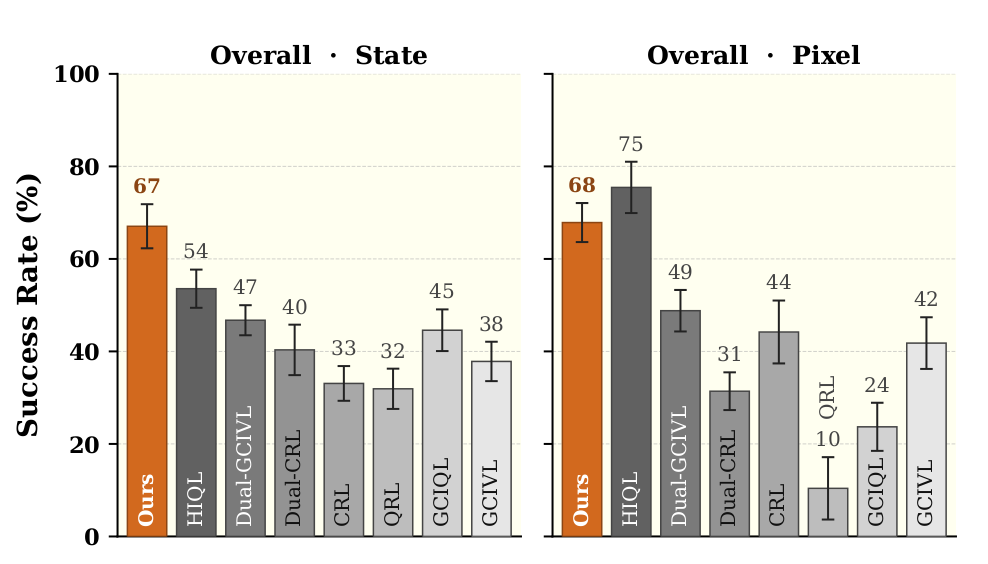}
    \caption{\label{fig:grouped_overview}\textbf{Overview.} Average success on 12 state-based and 5 pixel-based OGBench tasks spanning locomotion and manipulation. \ours{} achieves the best average on state-based tasks and is second only to the HIQL on pixel-based tasks.}
\end{figure}
\vspace{-3mm}

\cutsectionup
\section{Introduction}
\cutsectiondown

\label{sec:intro}
Advancements in representation learning have expanded the capabilities of deep reinforcement learning (RL). Self-supervised objectives \citep{schwarzer2021, srinivas2020curlcontrastiveunsupervisedrepresentations} and predictive world models \citep{Dreamerv1, tdmpc2} enable agents to capture underlying environment dynamics rather than memorizing trajectories \citep{schwarzer2021, srinivas2020curlcontrastiveunsupervisedrepresentations}. Yet deploying RL in real-world scenarios exposes a critical bottleneck: dense, task-specific rewards are notoriously difficult to design and collect. Agents must increasingly learn from offline datasets guided only by sparse, task-agnostic success signals.

Goal-conditioned RL (GCRL) provides a principled framework for this paradigm, tasking an agent to reach any specified goal from any starting state \citep{Kaelbling1993LearningTA, park2024ogbench}. Yet GCRL introduces profound representation challenges \citep{liu2022gcrlproblems}. To succeed, an agent must align its state and goal representations to understand not just what actions are physically possible, but how those actions contribute to maximizing the goal-conditioned cumulative reward. Under sparse rewards, standard methods lack the gradient feedback required to enforce this alignment, resulting in encoders that produce goal-agnostic features, leading to value overestimation, poor trajectory stitching, and brittle performance.

To learn effective goal-aware representations, we argue that following three key conditions can be beneficial. 1. \textit{Dynamical alignment} requires the encoder to capture immediate transition dynamics, grounding the agent in what actions are physically feasible. 2. \textit{Behavioral alignment} requires the encoder to organize state-goal pairs such that goal-directed actions and successor states are predictable from their latent representations. 3. \textit{Temporal alignment} requires the latent geometry to reflect goal-conditioned returns,

providing the critic with a pre-organized substrate rather than requiring it to discover this structure from sparse rewards alone.

Existing GCRL methods satisfy at most two of these conditions. Model-free methods such as GCBC \citep{gcbc1}, GCIVL \citep{gcivl}, and GCIQL \citep{kostrikov2021offlinereinforcementlearningimplicit} prioritize behavioral relations but bypass physical and temporal dynamics entirely. Representation-based methods such as Dual \citep{dual_goal} and VIP \citep{vip} target temporal alignment by leveraging value-based objectives and reward-distance estimation, respectively, where VIP specifically treats the distance between state and goal embeddings as a reward signal. However, they lack physical grounding and become brittle under high-dimensional observations or suboptimal data.

To address these limitations, we introduce \underline{M}ulti-\underline{s}cale \underline{P}redictive \underline{R}epresentations (\ours{}), a unified framework that jointly enforces all three alignment conditions. \ours{} operates at two complementary temporal granularities: at the local scale, it predicts immediate single-step transitions, grounding the encoder in physical dynamics; and at the global scale, it predicts goal-conditioned transitions and actions toward distant objectives, aligning the encoder with goal-directed intent. An end-to-end actor-critic agent trained on this structured representation benefits from a latent space where value approximation is substantially simplified. \ours{} is a flat, single-policy method: it introduces no subgoal generator, so hierarchical decomposition remains an orthogonal extension.
\vspace{-3pt}
\paragraph{Contribution.} We propose an end-to-end, multi-scale predictive framework that satisfies all three conditions. Unlike prior representation learning approaches that suffer from algorithmic brittleness \citep{dual_goal, byol,vip,alemi2019vib,TRA} , \ours{} constructs a robust latent space that maintains stability across diverse tasks. On the challenging OGBench benchmark, \ours{} achieves an average success rate of \textbf{67\%} on state-based and \textbf{68\%} on pixel-based tasks, improving over the prior representation learning baseline (Dual$_{\text{GCIVL}}$) by $~20$ points (Figure~\ref{fig:grouped_overview}). Notably, \ours{} outperforms the hierarchical state-of-the-art HIQL \citep{park2024hiqlofflinegoalconditionedrl} on state-based environments (54\%) without any hierarchy. \ours{} also remains effective when training data are scarce (Sec~\ref{sec:exp_robustness}),

\cutsectionup
\section{Related Work}
\cutsectiondown
\label{sec:related_work}
\textbf{Goal Representations for Offline GCRL.}
Representation learning in goal-conditioned RL (GCRL) seeks latent spaces that encode task progress and environment geometry~\citep{Kaelbling1993LearningTA, gcbc1}.
Prior approaches predominantly construct goal representations via temporal-distance metrics, contrastive objectives, or value-derived embeddings, such as Dual Goal Representations~\citep{dual_goal}, VIP~\citep{vip}, QRL~\citep{qrl}, HILP~\citep{park2024foundationpolicieshilbertrepresentations}, CRL~\citep{crl}, and state-abstraction methods~\citep{li2006towards,castro2019scalablemethodscomputingstate}.
However, because distance-based representations are coupled to or downstream of value estimation, errors in critic updates, particularly under high-dimensional observations, noisy offline data, and trajectory stitching, propagate directly back to the latent space, degrading downstream policy extraction \citep{lei2025qwsloptimizinggoalconditionedrl, yang2022rethinkinggoalconditionedsupervisedlearning}.
In contrast, \ours{} supervises the encoder with self-contained predictive targets decoupled from critic gradients, preventing critic error propagation and enabling representations to survive severe data degradation (Sec-\ref{sec:experiments}).

\textbf{Predictive Representations in RL.}
Predictive auxiliary tasks and latent world models~\citep{gelada2019deepmdplearningcontinuouslatent, schwarzer2021, fujimoto2025mrq, hafner2022deep, dreamerv3, tdmpc2, ha2018world, Dreamerv1, planet, Schrittwieser_2020, srinivas2020curlcontrastiveunsupervisedrepresentations} have established that forward-dynamics and reward prediction structure latent spaces effectively for single-task or dense-reward RL.
Leveraging system dynamics is a foundational approach for shaping representations in complex, partially observable environments \citep{litman2001, parr2008}. Moreover, existing latent-MDP guarantees (e.g., Theorem~1 \& 2 of MR.Q~\citep{fujimoto2025mrq}) require only transition and reward sufficiency for a single-task policy $\pi(a|z_s)$, whereas goal-conditioned control demands joint encoding of state-goal pairs $(s,g)$ such that goal-reaching actions remain recoverable. Proposition~\ref{prop:beh} states this goal-conditioned behavioral alignment condition.
Thus, rather than introducing new head architectures, our contribution establishes the predictive conditions required specifically for the sparse offline GCRL setting.

\cutsectionup
\section{Preliminary}
\cutsectiondown
\label{sec:preliminary}
\textbf{Offline Goal-Conditioned RL.}
We consider offline goal-conditioned reinforcement learning (GCRL) in an environment modeled as a goal-conditioned Markov decision process $\mathcal{M} = (\mathcal{S}, \mathcal{A}, \mathcal{P}, r, \gamma, \mathcal{G})$, where $\mathcal{S}$ is the state space, $\mathcal{A}$ is the action space, $\mathcal{P}(s'|s,a)$ describes transition dynamics, $\gamma \in (0, 1)$ is the discount factor, and $\mathcal{G} \subseteq \mathcal{S}$ is the goal space.
The agent learns from a static dataset of transition tuples $\mathcal{D} = \{(s, a, s', g)\}$, augmented via hindsight goal relabeling following standard OGBench protocols~\citep{park2024ogbench}.
The reward function $r(s, a, g)$ is sparse binary: $r(s,a,g) = 0$ if the state reaches goal $g$, and $-1$ otherwise. The objective is to learn a policy $\pi(a|s, g)$ that maximizes the expected cumulative discounted return $J(\pi) = \mathbb{E}_{\pi, \mathcal{P}}[\sum_{t=0}^{\infty} \gamma^t r(s_t, a_t, g)]$. In the rest of the manuscript, timestep subscripts are generally omitted to simplify the notation; instead, we use $(s, s')$ to indicate the current and next states.
To model long-horizon progress within trajectory rollouts, we define the empirical Monte-Carlo return along a dataset sequence as $r_{\text{MC}}(s, a, g) = \sum_{k=0}^{T-t} \gamma^k r(s_{t+k}, a_{t+k}, g)$. We write $Q^\beta(s,a,g) := \mathbb{E}_{\mathcal{D}}[\,r_{\text{MC}}(s,a,g) \mid s,a,g\,]$ for the conditional mean of the Monte-Carlo return under the relabeled dataset distribution; $r_{\text{MC}}$ is an unbiased sample of this quantity, which in general differs from the optimal value $Q^*$~\citep{Sutton:1988, quillen2018}.

\cutsectionup
\section{Method}
\cutsectiondown
\label{sec:method}
To construct a latent representation that captures both environment physics and long-horizon intent without relying on critic gradients, \ours{} jointly enforces the three alignment conditions introduced in Sec-\ref{sec:intro}.
We realize these conditions through four auxiliary predictive heads: a forward dynamics head $f_{\mathrm{dyn}}$ implementing \textit{dynamical alignment}, a return prediction head $f_{\mathrm{rew}}$ implementing \textit{temporal alignment}, and twin heads $f_{\mathrm{g\text{-}dyn}}$ and $f_{\mathrm{g\text{-}act}}$ implementing \textit{behavioral alignment}.
Each head corresponds to one of the three conditions. In the following subsections, we state how each objective controls the error for the downstream actor-critic, which only observes $(\mathbf{z}_s, \mathbf{z}_{sa}, \mathbf{z}_g)$ and not the raw observations. The complete set of encoders, predictive heads, and RL components is summarized in Figure~\ref{fig:pipeline} (left).

\noindent
\begin{minipage}[t]{0.48\linewidth}
\footnotesize
\centering
\vspace{0pt}
\phantomsection\label{tab:modules}
\vspace{-0.1cm}
\begin{tabular}{ll}
\toprule
\rowcolor{sb_orange!20}
\multicolumn{2}{c}{\textit{Encoders}} \\
State representation & $\mathbf{z}_s = \mathcal{E}^s_\psi(s)$ \\
State-action representation & $\mathbf{z}_{sa} = \mathcal{E}^{sa}_\psi(\mathbf{z}_s, a)$ \\
Goal representation & $\mathbf{z}_{g} = \mathcal{E}^s_\psi(g)$ \\
\rowcolor{sb_orange!20}
\multicolumn{2}{c}{\textit{Multi-scale Predictors}} \\
Forward dynamics & $\tilde{\mathbf{z}}_{s'} = f_{\mathrm{dyn}}(\mathbf{z}_{sa})$ \\
Goal-level dynamics & $\tilde{\mathbf{z}}^{\mathrm{goal}}_{s'} = f_{\mathrm{g\text{-}dyn}}(\mathbf{z}_s, \mathbf{z}_{g})$ \\
Goal-conditioned action & $\tilde{\mathbf{a}}^{\mathrm{goal}} = f_{\mathrm{g\text{-}act}}(\mathbf{z}_s, \mathbf{z}_{g})$ \\
Reward predictor & $\tilde{r} = f_{\mathrm{rew}}(\mathbf{z}_{sa},\, \mathbf{z}_{g})$ \\
\rowcolor{sb_orange!20}
\multicolumn{2}{c}{\textit{Goal-Conditioned RL}} \\
Critic function & $Q_\theta(\mathbf{z}_{sa}, \mathbf{z}_{g})$ \\
Actor policy & $\pi_\phi(a|\mathbf{z}_s, \mathbf{z}_{g})$ \\
\bottomrule
\end{tabular}
\end{minipage}%
\hfill
\begin{minipage}[t]{0.48\linewidth}
\vspace{5pt}
\centering
\includegraphics[width=\linewidth, trim=3mm 4mm 3mm 4mm, clip]{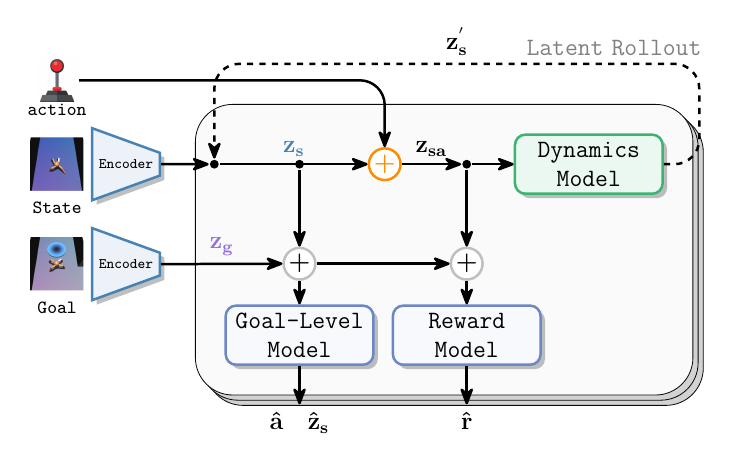}
\end{minipage}
\vspace{-1pt}
\captionof{figure}{\label{fig:pipeline}\textbf{Multi-scale Predictive Representations.} (\textit{Left}) Summary of \ours{} encoders, multi-scale predictors, and RL modules. (\textit{Right}) Architectural overview illustrating self-supervised predictive representation learning decoupled from actor-critic optimization.}
\vspace{6pt}

\paragraph{Population errors.}
Throughout, expectations are over $(s,a,s',g) \sim \mathcal{D}$ and $\mathbf{z}_{s'} = \mathcal{E}^s_{\bar\psi}(s')$ is the target-encoder representation. We work with expected values of the training losses; for the temporal head we take the best value achievable by \emph{any} head, so that the quantity depends on the encoder alone:
\begin{align}
  \delta_{\mathrm{dyn}}
    &= \E\big[\|f_{\mathrm{dyn}}(\mathbf{z}_{sa}) - \mathbf{z}_{s'}\|_2^2\big]
     \;=\; \E[\mathcal{L}_{\mathrm{dyn}}],
    \label{eq:delta_dyn}\\
  \Delta_{\mathrm{temp}}
    &= \min_{f}\ \E\big[\big(r_{\mathrm{MC}}(s,a,g) - f(\mathbf{z}_{sa}, \mathbf{z}_g)\big)^2\big]
     \;\le\; \E[\mathcal{L}_{\mathrm{rew}}],
    \label{eq:delta_temp}\\
  \delta_{\mathrm{beh}}
    &= \E\big[\|f_{\mathrm{g\text{-}act}}(\mathbf{z}_s, \mathbf{z}_g) - a\|_2^2\big]
     \;=\; \E[\mathcal{L}_{\mathrm{g\text{-}act}}].
    \label{eq:delta_beh}
\end{align}
The minimum in \eqref{eq:delta_temp} is over all measurable functions of $(\mathbf{z}_{sa}, \mathbf{z}_g)$; a trained $f_{\mathrm{rew}}$ attains at best this value, so reducing $\mathcal{L}_{\mathrm{rew}}$ reduces an upper bound on $\Delta_{\mathrm{temp}}$.

\subsection{Dynamical Alignment}
\label{sec:dyn_align}
\textit{Dynamical alignment} grounds the representation space with environment physics, ensuring that the encoder preserves feature necessary for forward simulation under arbitrary actions.
\paragraph{Forward Dynamics ($f_{\mathrm{dyn}}$).} Given the current state-action latent representation $\mathbf{z}_{sa} = \mathcal{E}^{sa}_\psi(\mathbf{z}_s, a)$, the forward dynamics module predicts the subsequent latent state $\tilde{\mathbf{z}}_{s'}$. Minimizing forward prediction error enforces transition consistency. The objective is formulated as:
\begin{equation}
\mathcal{L}_\mathrm{dyn} = \|\tilde{\mathbf{z}}_{s'} - \mathbf{z}_{s'}\|_2^2, \qquad \text{where} \quad \tilde{\mathbf{z}}_{s'} = f_{\mathrm{dyn}}(\mathbf{z}_{sa}).
\end{equation}
\textbf{Stability via Target Encoder.} Targets are computed by a target encoder $\mathcal{E}^s_{\bar{\psi}}$ with stopped gradients \citep{bardes2024revisitingfeaturepredictionlearning}, whose weights are copied from the online encoder every $K$ steps (Algorithm~\ref{alg:ours}). This ensures that the next-state latent is a stable regression target across updates, preventing the encoder from collapsing.  

\begin{proposition}[Latent transition consistency]
\label{prop:dyn}
Let $V : \mathcal{Z} \times \mathcal{Z} \to \R$ be any function that is $L_V$-Lipschitz in its first argument. Then
\begin{equation}
  \E\Big[\big(V(f_{\mathrm{dyn}}(\mathbf{z}_{sa}), \mathbf{z}_g) - V(\mathbf{z}_{s'}, \mathbf{z}_g)\big)^2\Big]
  \;\le\; L_V^2\, \delta_{\mathrm{dyn}}.
  \label{eq:prop_dyn}
\end{equation}
\end{proposition}
\noindent
Take $V$ to be the next-state value from which the critic bootstraps in Eq.~\eqref{eq:value_loss}. When $\delta_{\mathrm{dyn}}$ is small, $\mathbf{z}_{sa}$ retains the information needed to predict the value of the state it leads to, which is exactly what a Bellman-consistent critic on $(\mathbf{z}_{sa}, \mathbf{z}_g)$ must be able to represent. When it is large, two state-action pairs with different successors may share a latent, and no critic on that latent can assign them different values. Note that the critic does not use $f_{\mathrm{dyn}}$ directly; the proposition only says that $\mathbf{z}_{sa}$ keeps the information needed to predict the next state's value.

\subsection{Temporal Alignment}
\label{sec:temp_align}
While dynamical alignment models local transitions, it provides no information regarding the cumulative cost of reaching a goal, information the critic cannot recover from sparse binary rewards alone. \textit{Temporal alignment} addresses this by pre-organizing the latent space to reflect goal-conditioned return structure, reducing the burden on value estimation under sparse feedback. 

\paragraph{Return Predictor ($f_{\mathrm{rew}}$).} Given the state-action representation $\mathbf{z}_{sa}$ and goal embedding $\mathbf{z}_g$, the return prediction module outputs a scalar prediction $\tilde{r} = f_{\mathrm{rew}}(\mathbf{z}_{sa}, \mathbf{z}_g)$. Standard single-step reward targets carry no signal under sparse binary rewards ($r=0$ at goal, $-1$ elsewhere), remaining constant across almost all offline transitions. We address this by training $f_{\mathrm{rew}}$ to predict Monte-Carlo returns $r_{\text{MC}}$ computed over offline trajectories:
\begin{equation}
\mathcal{L}_\mathrm{rew} = \|\tilde{r} - r_{\mathrm{MC}}\|_2^2.
\end{equation}
Because $r_{\text{MC}}$ captures trajectory-level discounted returns, $\mathcal{L}_\mathrm{rew}$ provides a dense, multi-step regression signal that organizes latent states by goal proximity. Although $r_{\text{MC}}$ estimates the dataset return $Q^\beta$ (Sec-\ref{sec:preliminary}) rather than the optimal value, this dense ordering provides enough temporal structure to stabilize value learning, as the long-horizon locomotion results in Table~\ref{tab:ablation} show. \\

With $Q^\beta$ as defined in Sec-\ref{sec:preliminary}, let $\sigma^2_{\mathrm{MC}} := \E[(r_{\mathrm{MC}} - Q^\beta)^2]$ be the return variance that remains after conditioning on $(s,a,g)$, a property of the data independent of the encoder, and
$\varepsilon_V^{\min} := \min_{Q} \E\big[(Q(\mathbf{z}_{sa}, \mathbf{z}_g) - Q^\beta(s,a,g))^2\big]$
the smallest error that any critic on the latents can achieve against $Q^\beta$.
\begin{proposition}[Temporal alignment controls the representational value error]
\label{prop:temp}
\begin{equation}
  \Delta_{\mathrm{temp}} \;=\; \sigma^2_{\mathrm{MC}} \;+\; \varepsilon_V^{\min}.
  \label{eq:prop_temp}
\end{equation}
\end{proposition}
\noindent
Because $\sigma^2_{\mathrm{MC}}$ is fixed by the data, minimizing $\Delta_{\mathrm{temp}}$ over the encoder minimizes $\varepsilon_V^{\min}$, the value error that no critic, however expressive, can remove because the latent has discarded the information. The error against the \emph{optimal} value further splits into this representation term and a data term $\varepsilon_\beta$, the gap between $Q^\beta$ and $Q^*$, which no encoder can reduce (Corollary~\ref{cor:data_gap}). We expect $\varepsilon_\beta$ to be larger on datasets whose trajectories rarely reach the goal, such as the \texttt{stitch} datasets (Appendix~\ref{app:stitch}). The condition binds only for learned, compressive encoders (Remark~\ref{rem:scope}).

\subsection{Behavioral Alignment}
\label{sec:beh_align}
Dynamical and temporal alignment ground the encoder in physical transitions and goal proximity, but neither explicitly models the relationship between a current state and a target goal. \textit{Behavioral alignment} bridges this gap by supervising the latent space with goal-conditioned predictions, coupling state and goal representations directly to the control objective.

\textbf{Goal-level Dynamics ($f_{\mathrm{g\text{-}dyn}}$).} Given current state $\mathbf{z}_s$ and goal embedding $\mathbf{z}_g$, this module predicts the representation of the next state $\tilde{\mathbf{z}}_{s'}^{\mathrm{goal}}$ implied by goal-directed behavior. This forces $(\mathbf{z}_s, \mathbf{z}_g)$ to jointly encode the expected physical transition toward goal $g$:
\begin{equation}
\mathcal{L}_\mathrm{g\text{-}dyn} = \|\tilde{\mathbf{z}}_{s'}^{\mathrm{goal}} - \mathbf{z}_{s'}\|_2^2, \qquad \text{where} \quad \tilde{\mathbf{z}}_{s'}^{\mathrm{goal}} = f_{\mathrm{g\text{-}dyn}}(\mathbf{z}_s, \mathbf{z}_g).
\end{equation}

\textbf{Goal-conditioned Action Prediction ($f_{\mathrm{g\text{-}act}}$).} Given $(\mathbf{z}_s, \mathbf{z}_g)$, this module predicts the single-step dataset action $\tilde{\mathbf{a}}^{\mathrm{goal}}$ taken toward the goal:
\begin{equation}
\mathcal{L}_\mathrm{g\text{-}act} = \|\tilde{\mathbf{a}}^{\mathrm{goal}} - a\|_2^2, \qquad \text{where} \quad \tilde{\mathbf{a}}^{\mathrm{goal}} = f_{\mathrm{g\text{-}act}}(\mathbf{z}_s, \mathbf{z}_g).
\end{equation}
In contrast to local dynamics models that condition on actions $a$, $f_{\mathrm{g\text{-}dyn}}$ and $f_{\mathrm{g\text{-}act}}$ condition explicitly on goal $g$. This conditions state-goal pairs $(\mathbf{z}_s, \mathbf{z}_g)$ to encode goal-reaching behavior directly, establishing the behavioral alignment useful for effective policy extraction.\\

Let $\pi^*(s,g) \in \argmax_{a} Q^*(s,a,g)$ be an optimal action, assume $Q^*$ is $L_Q$-Lipschitz in its action argument, write $\hat{a} := f_{\mathrm{g\text{-}act}}(\mathbf{z}_s, \mathbf{z}_g)$, and let $\varepsilon_\pi := \E[\|a - \pi^*(s,g)\|_2^2]$ be the distance between dataset and optimal actions (a property of the data).
\begin{proposition}[Behavioral alignment bounds the one-step gap of a latent policy]
\label{prop:beh}
There exists a deterministic policy on the latent pair, namely $\hat{a}(\mathbf{z}_s, \mathbf{z}_g)$, whose expected one-step suboptimality satisfies
\begin{equation}
  \E\big[\,Q^*(s, \pi^*(s,g), g) - Q^*(s, \hat{a}, g)\,\big]
  \;\le\; L_Q \big(\sqrt{\delta_{\mathrm{beh}}} + \sqrt{\varepsilon_\pi}\,\big).
  \label{eq:prop_beh}
\end{equation}
\end{proposition}
\noindent
If $\delta_{\mathrm{beh}}$ is small, $(\mathbf{z}_s, \mathbf{z}_g)$ jointly encode enough of the state--goal relation for a near-optimal action to be read off them; the actor $\pi_\phi$ is trained by Eq.~\eqref{eq:actor}, not by copying $f_{\mathrm{g\text{-}act}}$, and the proposition says the latent has the capacity to support it. In single-task latent-MDP analyses~\citep{fujimoto2025mrq} the policy depends on $\mathbf{z}_s$ alone and no such condition arises; in the goal-conditioned setting $\pi^*$ depends jointly on $(s,g)$, and behavioral alignment is what makes that joint dependence representable. $f_{\mathrm{g\text{-}dyn}}$ introduces no fourth condition: its achievable error is bounded by $2\delta_{\mathrm{dyn}} + 2L_{\mathrm{dyn}}^2\delta_{\mathrm{beh}}$ (Remark~\ref{rem:gdyn}), but it supervises directly a quantity the other heads reach only by composition, which matters in practice (Sec-\ref{sec:ablations}).

\paragraph{The three terms are distinct.}
None of $\delta_{\mathrm{dyn}}$, $\Delta_{\mathrm{temp}}$, $\delta_{\mathrm{beh}}$ controls another: a latent can be transition-consistent yet discard goal distance, return can be predictable from $(\mathbf{z}_{sa}, \mathbf{z}_g)$ while $(\mathbf{z}_s, \mathbf{z}_g)$ does not determine the goal-directed action, and knowing that action does not make its consequence predictable (Appendix~\ref{app:distinct}). 

\subsection{Agent Training and Optimization}
\label{sec:training_objectives}

We jointly train the multi-scale representation modules and the downstream goal-conditioned policy using offline trajectory data. Optimization alternates between (i) multi-step representation updates on unrolled trajectory chunks and (ii) standard actor-critic updates on random transition batches. Algorithm~\ref{alg:ours} summarizes the procedure.

\paragraph{Representation Learning Objective.}
Given an unrolled trajectory chunk $\tau = \{(s_h, a_h, r_h, s_{h+1})\}_{h=0}^{H-1}$ sampled from $\mathcal{D}$ and goal $g$, we encode initial state and goal latents as $\mathbf{z}^0_s = \mathcal{E}^s_\psi(s_0)$ and $\mathbf{z}_g = \mathcal{E}^s_\psi(g)$. Unrolling latent dynamics over horizon $H$, we compute action-conditioned representations $\mathbf{z}^h_{sa} = \mathcal{E}^{sa}_\psi(\mathbf{z}^h_s, a_h)$ at step $h$. The total representation loss aggregates multi-scale prediction errors over horizon $H$:
\begin{equation}
\mathcal{L}_{\ours{}} = \sum_{h=0}^{H-1} \left[ \lambda_\mathrm{dyn} \mathcal{L}_\mathrm{dyn} + \lambda_\mathrm{rew} \mathcal{L}_\mathrm{rew} + \lambda_\mathrm{g\text{-}dyn} \mathcal{L}_\mathrm{g\text{-}dyn} + \lambda_\mathrm{g\text{-}act} \mathcal{L}_\mathrm{g\text{-}act} \right].
\label{eq:total_loss}
\end{equation}

\paragraph{Goal-Conditioned Actor-Critic.}
Value estimation updates two critics by minimizing Huber loss against a shared $n$-step TD target along dataset trajectory segments \citep{fujimoto2018addressing, fujimoto2025mrq}:
\begin{equation}
\mathcal{L}_Q = \sum_{i=1,2} \mathrm{Huber}_\delta\!\left( Q_{\theta_i}(\mathbf{z}_{sa}, \mathbf{z}_g) - \Big( R^{(n)}_t + \gamma^n \min_{j=1,2} Q_{\theta'_j}(\mathbf{z}^{n}_{sa}, \mathbf{z}_g) \Big) \right),
\label{eq:value_loss}
\end{equation}
where $R^{(n)}_t$ is the $n$-step discounted return, $\mathbf{z}^{n}_{sa}$ is the state-action representation $n$ steps ahead with the target policy's action, and $Q_{\theta'_j}$ are the target critics. The actor policy $\pi_\phi(a|\mathbf{z}_s, \mathbf{z}_g)$ is trained to maximize predicted value while constrained by behavior cloning regularization:
\begin{equation}
\mathcal{L}_\pi = - 0.5\sum_{i=1,2}Q_{\theta_i}(\mathbf{z}_{sa}, \mathbf{z}_g) + \lambda_{\mathrm{BC}} \|\pi_\phi(\mathbf{z}_s, \mathbf{z}_g) - a\|_2^2.
\label{eq:actor}
\end{equation}
Crucially, representation weights $\psi$ receive zero gradients from critic loss $\mathcal{L}_Q$ or actor loss $\mathcal{L}_\pi$. This actor-critic is the goal-conditioned TD3+BC agent~\citep{fujimoto2018addressing, fujimoto2021minimalist}; without the shared predictive encoder, we refer to it as GCTD3 and use it as the backbone baseline in Sec-\ref{sec:exp_attribution}.

\begin{algorithm}[h]
\caption{Representation and Policy Learning}
\label{alg:ours}
\begin{algorithmic}[1]
\STATE {\bfseries Input:} Offline dataset $\mathcal{D}$, horizon $H$, target update frequency $K$
\STATE {\bfseries Initialize:} Encoders $\psi$, predictors $\omega$, actor $\phi$, critic $\theta$; targets $\bar{\psi}, \bar{\phi}, \bar{\theta}$

\FOR{$t = 1 \dots T$}
    \IF{$t \bmod K = 0$}
        \STATE Update targets: $\bar{\psi} \leftarrow \psi,\; \bar{\phi} \leftarrow \phi,\; \bar{\theta} \leftarrow \theta$
        \FOR{$k = 1 \dots K$}
            \STATE Sample trajectory chunk $\tau \sim \mathcal{D}$
            \STATE Update representation $(\psi, \omega)$ by minimizing $\mathcal{L}_{\text{\ours{}}}(\tau)$ \hfill (Eq.~\ref{eq:total_loss})
        \ENDFOR
    \ENDIF

    \STATE Sample batch $(s,a,r,s',g) \sim \mathcal{D}$
    \STATE Update critic $\theta$ using TD loss $\mathcal{L}_Q$ \hfill (Eq.~\ref{eq:value_loss})
    \STATE Update actor $\phi$ using $\mathcal{L}_\pi$ \hfill (Eq.~\ref{eq:actor})

\ENDFOR
\end{algorithmic}
\end{algorithm}

\cutsectionup
\section{Experiments}
\cutsectionup
\label{sec:experiments}
Our experimental evaluation assesses the representations learned by \ours{} across state-based and pixel-based offline goal-conditioned reinforcement learning (GCRL) tasks from OGBench~\citep{park2024ogbench}. After reporting the main benchmark results, we structure our analysis around four core questions:
\begin{enumerate}
    \item[\textbf{(RQ1)}] \textbf{Ablation:} Which alignment condition matters most, and where?
    \item[\textbf{(RQ2)}] \textbf{Backbone:} How much of the gain comes from the representation?
    \item[\textbf{(RQ3)}] \textbf{Data efficiency:} Does the predictive representation help when data is limited?
    \item[\textbf{(RQ4)}] \textbf{Failure modes:} Where does \ours{} fall short, and why?
\end{enumerate}

\paragraph{Tasks and Environments.}
For main benchmark, we evaluate across 12 state-based and 5 pixel-based environments from OGBench~\citep{park2024ogbench}, covering two main domains: \textit{locomotion} (AntMaze, PointMaze, HumanoidMaze) and \textit{manipulation} (Cube, Scene, Puzzle), using the \texttt{navigate} and \texttt{play} dataset splits. Additional results on the \texttt{stitch} datasets are reported in Appendix~\ref{app:stitch}.

\paragraph{Baselines.}
We compare against standard GCRL baselines: \textbf{GCBC}~\citep{gcbc1}, \textbf{GCIVL}~\citep{gcivl}, \textbf{GCIQL}~\citep{kostrikov2021offlinereinforcementlearningimplicit}, \textbf{QRL}~\citep{qrl}, \textbf{CRL}~\citep{crl}, and the hierarchical \textbf{HIQL}~\citep{park2024hiqlofflinegoalconditionedrl}. As the primary goal-representation benchmark, we evaluate \textbf{Dual Goal Representations}~\citep{dual_goal} on the two best backbones reported, GCIVL and CRL ($\text{Dual}_\text{GCIVL}$, $\text{Dual}_\text{CRL}$). Additionally, to isolate the effect of representation learning, we compare against \textbf{GCTD3}~\citep{fujimoto2021minimalist}, the TD3+BC agent that \ours{} is built on. GCTD3 follows the OGBench network design~\citep{park2024ogbench}: the actor and critic each learn their own features from their own losses. \ours{} differs only in the encoder, which is shared between the actor and critic and trained using the predictive losses.

\paragraph{Target Horizon ($n$-step).}
Since $n$-step TD targets significantly influence offline GCRL performance, we re-run all TD-based baselines (GCIQL, GCIVL, $\text{Dual}_\text{GCIVL}$, and HIQL) using $n=3$; results at $n=1$ are reported in Appendix~\ref{app:qh1_results}. $n$-step targets are undefined for GCBC, QRL, CRL, and $\text{Dual}_\text{CRL}$, which are reported using published numbers.

\paragraph{Evaluation.}
State-based results are averaged over 8 random seeds and pixel-based results over 4, reporting the mean success rate across the final three checkpoints. For pixel-based OGBench~\citep{park2024ogbench}, prior work reported that frame stacking (FS) does not improve performance; we therefore use single frames for baselines. In contrast, \ours{} uses three stacked frames. We isolate the specific effect of FS on both \ours{} and GCTD3 in Sec-\ref{sec:ablations}.

\subsection{Main Results}
\label{sec:exp_main}

\begin{table*}[t]
\centering
\begin{adjustbox}{width=0.95\textwidth,keepaspectratio}
\begin{tabular}{l *{5}{c}  *{3}{c} c}
\toprule
\textbf{Environment}
  & \textbf{GCBC} & \textbf{GCIVL} & \textbf{GCIQL} & \textbf{QRL} & \textbf{CRL}
  & $\textbf{\small Dual}_{\textbf{\tiny GCIVL}}$ & $\textbf{\small Dual}_{\textbf{\tiny CRL}}$
  & \textbf{HIQL}
  & \textbf{OURS} \\
\midrule
\rowcolor{sb_orange!20}\multicolumn{10}{c}{\textbf{ State-based environments}} \\
\texttt{pointmaze-medium-navigate}
  & $9$ {\tiny $\pm 6$} & $79$ {\tiny $\pm 1$} & $66$ {\tiny $\pm 8$} & $\mathbf{82}$ {\tiny $\pm 5$} & $29$ {\tiny $\pm 7$}
  & $\mathbf{86}$ {\tiny $\pm 2$} & $33$ {\tiny $\pm 10$}
  & $79$ {\tiny $\pm 3$}
  & $54$ {\tiny $\pm 6$} \\
\texttt{pointmaze-large-navigate}
  & $29$ {\tiny $\pm 6$} & $65$ {\tiny $\pm 8$} & $38$ {\tiny $\pm 2$} & $\mathbf{86}$ {\tiny $\pm 9$} & $39$ {\tiny $\pm 7$}
  & $67$ {\tiny $\pm 4$} & $39$ {\tiny $\pm 12$}
  & $73$ {\tiny $\pm 5$}
  & $48$ {\tiny $\pm 8$} \\
\texttt{antmaze-medium-navigate}
  & $29$ {\tiny $\pm 4$} & $87$ {\tiny $\pm 1$} & $90$ {\tiny $\pm 3$} & $88$ {\tiny $\pm 3$} & $\mathbf{95}$ {\tiny $\pm 1$}
  & $82$ {\tiny $\pm 4$} & $93$ {\tiny $\pm 3$}
  & $\mathbf{96}$ {\tiny $\pm 2$}
  & $\mathbf{99}$ {\tiny $\pm 1$} \\
\texttt{antmaze-large-navigate}
  & $24$ {\tiny $\pm 2$} & $33$ {\tiny $\pm 7$} & $40$ {\tiny $\pm 5$} & $75$ {\tiny $\pm 6$} & $83$ {\tiny $\pm 4$}
  & $41$ {\tiny $\pm 1$} & $87$ {\tiny $\pm 2$}
  & $92$ {\tiny $\pm 2$}
  & $\mathbf{98}$ {\tiny $\pm 2$} \\
\texttt{antmaze-giant-navigate}
  & $0$ {\tiny $\pm 0$} & $1$ {\tiny $\pm 1$} & $1$ {\tiny $\pm 1$} & $14$ {\tiny $\pm 3$} & $16$ {\tiny $\pm 3$}
  & $1$ {\tiny $\pm 2$} & $21$ {\tiny $\pm 4$}
  & $\mathbf{68}$ {\tiny $\pm 6$}
  & $\mathbf{68}$ {\tiny $\pm 7$} \\
\texttt{humanoidmaze-medium-navigate}
  & $8$ {\tiny $\pm 2$} & $31$ {\tiny $\pm 4$} & $34$ {\tiny $\pm 3$} & $21$ {\tiny $\pm 8$} & $60$ {\tiny $\pm 4$}
  & $32$ {\tiny $\pm 4$} & $57$ {\tiny $\pm 4$}
  & $\mathbf{92}$ {\tiny $\pm 2$}
  & $67$ {\tiny $\pm 5$} \\
\texttt{humanoidmaze-large-navigate}
  & $1$ {\tiny $\pm 0$} & $3$ {\tiny $\pm 1$} & $7$ {\tiny $\pm 3$} & $5$ {\tiny $\pm 1$} & $24$ {\tiny $\pm 4$}
  & $5$ {\tiny $\pm 2$} & $18$ {\tiny $\pm 4$}
  & $\mathbf{58}$ {\tiny $\pm 3$}
  & $\mathbf{58}$ {\tiny $\pm 3$} \\
\texttt{cube-single-play}
  & $6$ {\tiny $\pm 2$} & $53$ {\tiny $\pm 6$} & $60$ {\tiny $\pm 3$} & $5$ {\tiny $\pm 1$} & $19$ {\tiny $\pm 2$}
  & $\mathbf{80}$ {\tiny $\pm 3$} & $60$ {\tiny $\pm 1$}
  & $16$ {\tiny $\pm 4$}
  & $65$ {\tiny $\pm 5$} \\
\texttt{cube-double-play}
  & $1$ {\tiny $\pm 1$} & $36$ {\tiny $\pm 3$} & $42$ {\tiny $\pm 2$} & $1$ {\tiny $\pm 0$} & $10$ {\tiny $\pm 2$}
  & $\mathbf{57}$ {\tiny $\pm 6$} & $24$ {\tiny $\pm 5$}
  & $6$ {\tiny $\pm 2$}
  & $13$ {\tiny $\pm 3$} \\
\texttt{scene-play}
  & $5$ {\tiny $\pm 1$} & $55$ {\tiny $\pm 5$} & $35$ {\tiny $\pm 5$} & $5$ {\tiny $\pm 1$} & $19$ {\tiny $\pm 2$}
  & $\mathbf{91}$ {\tiny $\pm 4$} & $44$ {\tiny $\pm 5$}
  & $43$ {\tiny $\pm 9$}
  & $45$ {\tiny $\pm 5$} \\
\texttt{puzzle-3x3-play}
  & $2$ {\tiny $\pm 0$} & $3$ {\tiny $\pm 2$} & $90$ {\tiny $\pm 8$} & $1$ {\tiny $\pm 0$} & $3$ {\tiny $\pm 1$}
  & $5$ {\tiny $\pm 1$} & $6$ {\tiny $\pm 1$}
  & $13$ {\tiny $\pm 3$}
  & $\mathbf{98}$ {\tiny $\pm 1$} \\
\texttt{puzzle-4x4-play}
  & $0$ {\tiny $\pm 0$} & $8$ {\tiny $\pm 4$} & $33$ {\tiny $\pm 5$} & $0$ {\tiny $\pm 0$} & $0$ {\tiny $\pm 0$}
  & $14$ {\tiny $\pm 3$} & $2$ {\tiny $\pm 0$}
  & $7$ {\tiny $\pm 2$}
  & $\mathbf{93}$ {\tiny $\pm 5$} \\
\midrule
\textbf{Avg. State}
  & $10$ {\tiny $\pm 1$} & $38$ {\tiny $\pm 1$} & $45$ {\tiny $\pm 1$} & $32$ {\tiny $\pm 1$} & $33$ {\tiny $\pm 1$}
  & $47$ {\tiny $\pm 1$} & $40$ {\tiny $\pm 2$}
  & $54$ {\tiny $\pm 1$}
  & $\mathbf{67}$ {\tiny $\pm 1$} \\
\midrule
\rowcolor{sb_orange!20}\multicolumn{10}{c}{\textbf{Pixel-based environments}} \\
\vamn
  & $11$ {\tiny $\pm 2$} & $48$ {\tiny $\pm 1$} & $36$ {\tiny $\pm 6$} & $0$ {\tiny $\pm 0$} & $\mathbf{95}$ {\tiny $\pm 1$}
  & $85$ {\tiny $\pm 2$} & $85$ {\tiny $\pm 3$}
  & $\mathbf{94}$ {\tiny $\pm 2$}
  & $90$ {\tiny $\pm 3$} \\
\valn
  & $4$ {\tiny $\pm 0$} & $25$ {\tiny $\pm 8$} & $7$ {\tiny $\pm 3$} & $0$ {\tiny $\pm 0$} & $\mathbf{84}$ {\tiny $\pm 1$}
  & $57$ {\tiny $\pm 3$} & $43$ {\tiny $\pm 3$}
  & $\mathbf{81}$ {\tiny $\pm 1$}
  & $\mathbf{81}$ {\tiny $\pm 4$} \\
\vcsp
  & $5$ {\tiny $\pm 1$} & $65$ {\tiny $\pm 8$} & $50$ {\tiny $\pm 9$} & $41$ {\tiny $\pm 15$} & $31$ {\tiny $\pm 15$}
  & $64$ {\tiny $\pm 6$} & $14$ {\tiny $\pm 8$}
  & $\mathbf{81}$ {\tiny $\pm 4$}
  & $\mathbf{82}$ {\tiny $\pm 5$} \\
\vsp
  & $12$ {\tiny $\pm 2$} & $33$ {\tiny $\pm 2$} & $15$ {\tiny $\pm 1$} & $10$ {\tiny $\pm 1$} & $11$ {\tiny $\pm 2$}
  & $38$ {\tiny $\pm 7$} & $15$ {\tiny $\pm 1$}
  & $46$ {\tiny $\pm 2$}
  & $\mathbf{56}$ {\tiny $\pm 5$} \\
\vptp
  & $0$ {\tiny $\pm 0$} & $38$ {\tiny $\pm 4$} & $10$ {\tiny $\pm 2$} & $1$ {\tiny $\pm 1$} & $0$ {\tiny $\pm 0$}
  & $0$ {\tiny $\pm 0$} & $0$ {\tiny $\pm 0$}
  & $\mathbf{75}$ {\tiny $\pm 11$}
  & $30$ {\tiny $\pm 4$} \\
\midrule
\textbf{Avg. Pixel}
  & $6$ {\tiny $\pm 1$} & $42$ {\tiny $\pm 2$} & $24$ {\tiny $\pm 2$} & $10$ {\tiny $\pm 3$} & $44$ {\tiny $\pm 3$}
  & $49$ {\tiny $\pm 2$} & $31$ {\tiny $\pm 2$}
  & $\mathbf{75}$ {\tiny $\pm 2$}
  & $68$ {\tiny $\pm 2$} \\
\bottomrule
\end{tabular}
\end{adjustbox}

\vspace{-3pt}
\caption{\label{tab:full_results}\textbf{Main Results} across state-based and pixel-based settings. Bold: within 95\% of the row best. GCBC, QRL, CRL, and $\text{Dual}_\text{CRL}$: published numbers; all others re-run with $n=3$.}
\end{table*}

Table~\ref{tab:full_results} shows that \ours{} leads on average across state-based tasks (67 versus 54 for HIQL), while HIQL leads on pixel-based tasks (75 versus 68). Much of the gap on pixel-based tasks comes from \texttt{visual-puzzle-3x3}, where \ours{} scores 30 compared with HIQL's 75. While other baselines are stronger in a particular category, either manipulation or locomotion, our method performs well on environments from both categories. \ours{} is particularly strong on \texttt{antmaze} and on the state-based puzzle tasks, where it achieves the best results. On pixel-based tasks, it is the best or within 5 points of the best on four of the five environments, and falls clearly behind only on \texttt{visual-puzzle-3x3}. \\

The performance on \texttt{PointMaze} is more surprising. Despite appearing simpler than the other locomotion tasks, these environments remain difficult for our method. The drop from \texttt{cube-single} (65) to \texttt{cube-double} (13) is also striking. This weakness might partly come from the TD3's Actor-Critc backbone (Appendix Table~\ref{tab:td3}), and the ablations (Sec-\ref{sec:ablations}) show that no single component is specifically hurting the performance. One possible explanation is that handling two cubes requires more complex decision making that the representation does not reliably preserve. The additional object may also make the goal-conditioned representation harder to learn.

\subsection{Ablations and Analysis}
\label{sec:ablations}

\begin{table}[t]
\centering
\begin{adjustbox}{width=0.8\textwidth,keepaspectratio}
\begin{tabular}{l *{6}{c} c}
\toprule
\textbf{Environment}
  & \textbf{w/o} $\boldsymbol{\mathcal{L}_{\mathrm{dyn}}}$
  & \textbf{w/o} $\boldsymbol{\mathcal{L}_{\mathrm{rew}}}$
  & \textbf{w/o} $\boldsymbol{\mathcal{L}_{\mathrm{g\text{-}dyn}}}$
  & \textbf{w/o} $\boldsymbol{\mathcal{L}_{\mathrm{g\text{-}act}}}$
  & $\boldsymbol{n\!=\!1}$
  & \textbf{w/o FS}
  & \textbf{OURS} \\
\midrule
\rowcolor{sb_orange!30}\multicolumn{8}{c}{\textit{Pixel-based environments}} \\
\vamn
  & \cellcolor{red!5}$89$ {\tiny $\pm 4$}
  & \cellcolor{red!5}$89$ {\tiny $\pm 3$}
  & \cellcolor{red!5}$87$ {\tiny $\pm 2$}
  & $90$ {\tiny $\pm 3$}
  & \cellcolor{green!5}$92$ {\tiny $\pm 3$}
  & \cellcolor{red!5}$86$ {\tiny $\pm 4$}
  & $90$ {\tiny $\pm 3$} \\
\valn
  & \cellcolor{red!15}$72$ {\tiny $\pm 7$}
  & \cellcolor{red!30}$28$ {\tiny $\pm 10$}
  & \cellcolor{red!15}$69$ {\tiny $\pm 5$}
  & \cellcolor{green!15}$86$ {\tiny $\pm 3$}
  & \cellcolor{red!5}$77$ {\tiny $\pm 5$}
  & \cellcolor{red!5}$78$ {\tiny $\pm 5$}
  & $81$ {\tiny $\pm 4$} \\
\vcsp
  & \cellcolor{green!5}$83$ {\tiny $\pm 5$}
  & \cellcolor{red!15}$71$ {\tiny $\pm 3$}
  & \cellcolor{green!15}$87$ {\tiny $\pm 1$}
  & \cellcolor{red!5}$79$ {\tiny $\pm 9$}
  & $82$ {\tiny $\pm 2$}
  & \cellcolor{red!30}$63$ {\tiny $\pm 1$}
  & $82$ {\tiny $\pm 5$} \\
\vsp
  & \cellcolor{green!5}$57$ {\tiny $\pm 7$}
  & \cellcolor{green!15}$65$ {\tiny $\pm 2$}
  & \cellcolor{red!30}$32$ {\tiny $\pm 2$}
  & \cellcolor{red!15}$49$ {\tiny $\pm 6$}
  & \cellcolor{red!30}$36$ {\tiny $\pm 6$}
  & \cellcolor{red!30}$25$ {\tiny $\pm 3$}
  & $56$ {\tiny $\pm 5$} \\
\vptp
  & \cellcolor{red!15}$21$ {\tiny $\pm 1$}
  & \cellcolor{green!15}$39$ {\tiny $\pm 2$}
  & \cellcolor{red!30}$0$ {\tiny $\pm 0$}
  & \cellcolor{red!15}$25$ {\tiny $\pm 2$}
  & \cellcolor{red!15}$20$ {\tiny $\pm 1$}
  & \cellcolor{red!15}$21$ {\tiny $\pm 1$}
  & $30$ {\tiny $\pm 4$} \\
\midrule
\textbf{Average}
  & \cellcolor{red!5}$64$ {\tiny $\pm 2$}
  & \cellcolor{red!15}$58$ {\tiny $\pm 2$}
  & \cellcolor{red!15}$55$ {\tiny $\pm 1$}
  & \cellcolor{red!5}$66$ {\tiny $\pm 2$}
  & \cellcolor{red!15}$61$ {\tiny $\pm 2$}
  & \cellcolor{red!15}$55$ {\tiny $\pm 1$}
  & $\mathbf{68}$ {\tiny $\pm 2$} \\
\midrule
\rowcolor{sb_orange!30}\multicolumn{8}{c}{\textit{State-based environments}} \\
\texttt{pointmaze-medium-navigate}
  & \cellcolor{red!15}$43$ {\tiny $\pm 4$}
  & \cellcolor{red!30}$36$ {\tiny $\pm 7$}
  & \cellcolor{green!15}$60$ {\tiny $\pm 3$}
  & \cellcolor{red!15}$47$ {\tiny $\pm 6$}
  & \cellcolor{red!5}$50$ {\tiny $\pm 1$}
  & --
  & $54$ {\tiny $\pm 6$} \\
\texttt{pointmaze-large-navigate}
  & \cellcolor{red!30}$24$ {\tiny $\pm 9$}
  & \cellcolor{red!30}$31$ {\tiny $\pm 3$}
  & \cellcolor{green!15}$62$ {\tiny $\pm 3$}
  & \cellcolor{red!30}$33$ {\tiny $\pm 3$}
  & \cellcolor{red!30}$30$ {\tiny $\pm 2$}
  & --
  & $48$ {\tiny $\pm 8$} \\
\texttt{antmaze-medium-navigate}
  & \cellcolor{red!5}$96$ {\tiny $\pm 2$}
  & \cellcolor{red!5}$98$ {\tiny $\pm 1$}
  & $99$ {\tiny $\pm 0$}
  & \cellcolor{red!5}$98$ {\tiny $\pm 0$}
  & \cellcolor{red!5}$98$ {\tiny $\pm 1$}
  & --
  & $\mathbf{99}$ {\tiny $\pm 1$} \\
\texttt{antmaze-large-navigate}
  & \cellcolor{red!5}$97$ {\tiny $\pm 1$}
  & \cellcolor{red!15}$92$ {\tiny $\pm 3$}
  & $98$ {\tiny $\pm 1$}
  & \cellcolor{red!5}$97$ {\tiny $\pm 1$}
  & \cellcolor{red!5}$96$ {\tiny $\pm 1$}
  & --
  & $\mathbf{98}$ {\tiny $\pm 2$} \\
\texttt{antmaze-giant-navigate}
  & \cellcolor{red!15}$54$ {\tiny $\pm 1$}
  & \cellcolor{red!30}$0$ {\tiny $\pm 0$}
  & \cellcolor{red!15}$57$ {\tiny $\pm 1$}
  & \cellcolor{red!15}$55$ {\tiny $\pm 3$}
  & \cellcolor{red!30}$43$ {\tiny $\pm 4$}
  & --
  & $\mathbf{68}$ {\tiny $\pm 7$} \\
\texttt{humanoidmaze-medium-navigate}
  & \cellcolor{red!15}$56$ {\tiny $\pm 3$}
  & \cellcolor{red!30}$0$ {\tiny $\pm 0$}
  & \cellcolor{red!30}$0$ {\tiny $\pm 0$}
  & \cellcolor{green!15}$75$ {\tiny $\pm 9$}
  & \cellcolor{red!30}$35$ {\tiny $\pm 3$}
  & --
  & $67$ {\tiny $\pm 5$} \\
\texttt{humanoidmaze-large-navigate}
  & \cellcolor{red!15}$53$ {\tiny $\pm 2$}
  & \cellcolor{red!30}$0$ {\tiny $\pm 0$}
  & \cellcolor{red!30}$0$ {\tiny $\pm 0$}
  & \cellcolor{red!30}$43$ {\tiny $\pm 5$}
  & \cellcolor{red!30}$11$ {\tiny $\pm 3$}
  & --
  & $\mathbf{58}$ {\tiny $\pm 3$} \\
\texttt{cube-single-play}
  & \cellcolor{red!30}$50$ {\tiny $\pm 3$}
  & \cellcolor{red!5}$63$ {\tiny $\pm 2$}
  & \cellcolor{green!5}$68$ {\tiny $\pm 6$}
  & \cellcolor{red!5}$62$ {\tiny $\pm 4$}
  & \cellcolor{red!30}$46$ {\tiny $\pm 4$}
  & --
  & $65$ {\tiny $\pm 5$} \\
\texttt{cube-double-play}
  & \cellcolor{green!5}$14$ {\tiny $\pm 2$}
  & \cellcolor{red!5}$12$ {\tiny $\pm 4$}
  & \cellcolor{red!5}$12$ {\tiny $\pm 1$}
  & \cellcolor{red!5}$11$ {\tiny $\pm 2$}
  & \cellcolor{red!5}$11$ {\tiny $\pm 2$}
  & --
  & $13$ {\tiny $\pm 3$} \\
\texttt{scene-play}
  & $45$ {\tiny $\pm 2$}
  & \cellcolor{green!5}$47$ {\tiny $\pm 1$}
  & \cellcolor{red!5}$44$ {\tiny $\pm 4$}
  & \cellcolor{red!15}$38$ {\tiny $\pm 1$}
  & \cellcolor{red!30}$30$ {\tiny $\pm 4$}
  & --
  & $45$ {\tiny $\pm 5$} \\
\texttt{puzzle-3x3-play}
  & \cellcolor{red!15}$93$ {\tiny $\pm 1$}
  & \cellcolor{green!5}$99$ {\tiny $\pm 0$}
  & \cellcolor{green!5}$99$ {\tiny $\pm 0$}
  & \cellcolor{green!5}$100$ {\tiny $\pm 1$}
  & \cellcolor{red!30}$81$ {\tiny $\pm 4$}
  & --
  & $98$ {\tiny $\pm 1$} \\
\texttt{puzzle-4x4-play}
  & \cellcolor{red!15}$81$ {\tiny $\pm 3$}
  & \cellcolor{red!15}$84$ {\tiny $\pm 5$}
  & \cellcolor{green!5}$95$ {\tiny $\pm 0$}
  & \cellcolor{green!5}$94$ {\tiny $\pm 1$}
  & \cellcolor{red!30}$73$ {\tiny $\pm 2$}
  & --
  & $93$ {\tiny $\pm 5$} \\
\midrule
\textbf{Average}
  & \cellcolor{red!15}$59$ {\tiny $\pm 1$}
  & \cellcolor{red!30}$47$ {\tiny $\pm 1$}
  & \cellcolor{red!15}$58$ {\tiny $\pm 1$}
  & \cellcolor{red!5}$63$ {\tiny $\pm 1$}
  & \cellcolor{red!30}$50$ {\tiny $\pm 1$}
  & --
  & $\mathbf{67}$ {\tiny $\pm 1$} \\
\bottomrule
\end{tabular}
\end{adjustbox}

\vspace{-2mm}
\caption{\label{tab:ablation}\textbf{Ablation study.} Each cell shows the variant's score. Red (\colorbox{red!30}{\phantom{x}}\,\colorbox{red!15}{\phantom{x}}\,\colorbox{red!5}{\phantom{x}}) and green (\colorbox{green!30}{\phantom{x}}\,\colorbox{green!15}{\phantom{x}}\,\colorbox{green!5}{\phantom{x}}) indicate performance w.r.t \ours{}, respectively, by $\geq\!15$, $[5,15)$, and $[0,5)$ success rate. }
\vspace{-6mm}
\end{table}

To understand how each alignment objective contributes to the overall performance, we perform a leave-one-out ablation study. Additionally, we also study the contribution of n-step prediction and frame stacking (Pixel envs only) with respect to the full version of MSPR. These ablations help us understand which components are important and identify the environments where our method still falls behind other approaches.

\paragraph{Condition sensitivity (RQ1).}Given that these are goal-conditioned tasks, removing the temporal alignment objective $\mathcal{L}_{\mathrm{rew}}$ has a clear impact on performance. Manipulation environments are less affected, while the impact is much larger on locomotion, especially for long-horizon tasks such as \texttt{antmaze-giant}, \texttt{humanoidmaze-medium}, and \texttt{humanoidmaze-large}, where performance drops to zero. Two exceptions are \texttt{visual-puzzle-3x3} and \texttt{visual-scene}, which perform better without temporal alignment.  
Removing $\mathcal{L}_{\mathrm{g\text{-}dyn}}$ reduces the average performance from 67 to 58 and causes a similar collapse on the two humanoid maze tasks as removing temporal alignment.\\

On \texttt{visual-puzzle-3x3}, performance also drops to zero, suggesting that behavioral alignment, particularly the goal-conditioned dynamics objective, helps learn representations that are important for this task. In comparison, $\mathcal{L}_{\mathrm{g\text{-}act}}$ (Proposition~\ref{prop:beh}) has the smallest effect on average, although removing it still leads to a 15-point drop on \texttt{humanoidmaze-large}. Finally, dropping $\mathcal{L}_{\mathrm{dyn}}$ (Proposition~\ref{prop:dyn}) costs 8 points on state and 4 on pixel, and the drop is spread across environments rather than concentrated. Frame stacking, in particular, provides a clear improvement on visual manipulation tasks, while the improvements on locomotion tasks are relatively smaller.  Finally, as we scale \ours{} to longer-horizon tasks, n-step returns become important, with a larger effect on state-based environments. Removing any single objective does not outperform the full \ours{}, but the ablations show how each objective contributes to the overall performance.

\begin{figure}[h]
    \centering
    \includegraphics[width=0.95\linewidth, trim=2mm 2mm 2mm 2mm, clip]{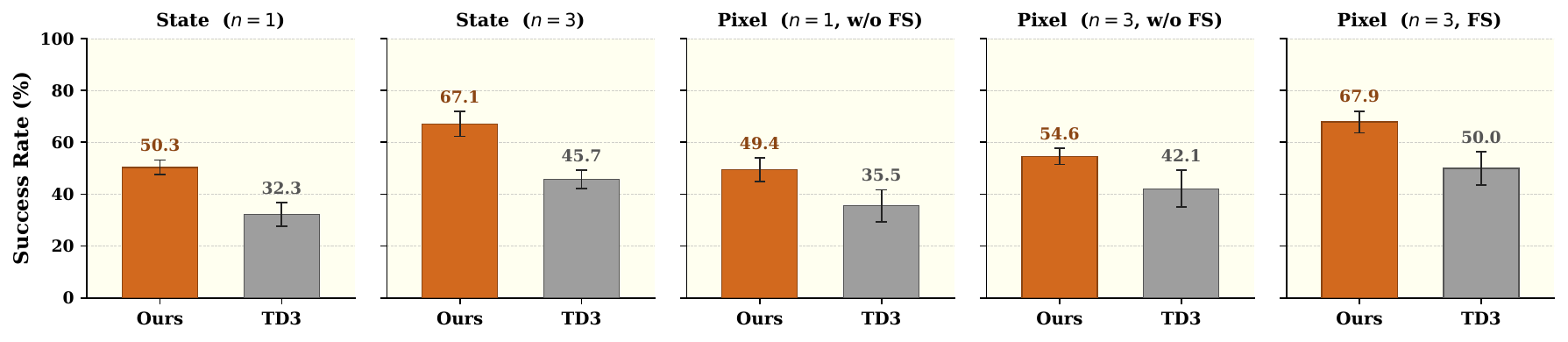}
    \vspace{-3pt}
    \caption{\label{fig:td3_backbone}\textbf{Representation vs.\ backbone.} Average success of \ours{} (orange) and its GCTD3 backbone (grey). Both use the same TD3+BC agent and $n$-step targets. Per-environment numbers in Appendix Tables~\ref{tab:td3} and~\ref{tab:td3_visual}.}
    \vspace{-3pt}
\end{figure}

\paragraph{Representation vs.\ backbone (RQ2).}
\label{sec:exp_attribution}
Following the same architecture as the OGBench baselines, GCTD3 shares the actor-critic algorithm and $n$-step targets with \ours{}. The only difference is the representation learning used in \ours{}, allowing us to isolate the gains from the predictive representation. Each subplot in Figure~\ref{fig:td3_backbone} compares the two agents under the same setting, and in every case \ours{} performs better, showing the positive impact of learning representations through the alignment conditions. Across subplots, both agents improve by a similar amount (17 and 14 points on state), suggesting that the gain from $n$-step targets mainly comes from the TD3+BC backbone. This also suggests that the comparison is fair, as at $n\!=\!3$ the backbone reaches 46 on state, close to the other baselines. This is consistent with Proposition~\ref{prop:temp}: The predictive losses, in particular return prediction, organize the latent space by proximity to the goal, helping the critic learn a value function that reflects this proximity. The Q-value traces in Appendix~\ref{sec:failure-case} illustrate this, with the critic's value increasing as the agent approaches the goal (Figure~\ref{fig:valn_success}). Per-environment results are provided in Appendix Tables~\ref{tab:td3} and~\ref{tab:td3_visual}.

\paragraph{Frame stacking on pixel-based tasks.} OGBench reports that frame stacking does not help the baselines on pixel-based tasks \citep{park2024ogbench}, so we follow that setup and run the baselines with single frames. For our method, we isolate its effect by running both \ours{} and its GCTD3 backbone with and without frame stacking. Frame stacking helps both, but it helps \ours{} more, and the gain is largest on manipulation tasks, where the agent must infer object motion from consecutive frames. Without frame stacking, \ours{} still clearly improves on GCTD3, isolating the gain from predictive representation.

\begin{figure}[h]
    \centering
    \includegraphics[width=0.95\linewidth, trim=2mm 2mm 2mm 2mm, clip]{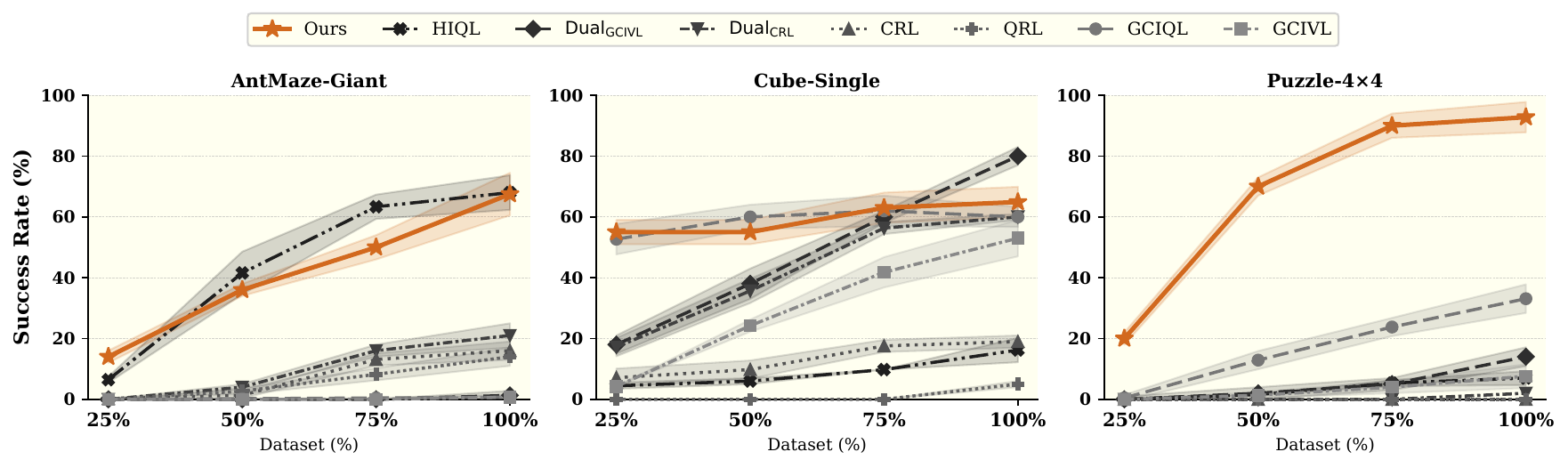}
    \vspace{-2mm}
    \caption{\label{fig:data_efficiency}\textbf{Data efficiency} under reduced training data ($25\%$, $50\%$, $75\%$, $100\%$); shaded bands are one std over 4 seeds.}
    \vspace{-2mm}
\end{figure}

\paragraph{Data Efficiency (RQ3).}\label{sec:exp_robustness} To test how much the predictive representation helps when data is limited, we train each method on $25\%$, $50\%$, $75\%$, and $100\%$ of the original offline datasets (Figure~\ref{fig:data_efficiency}). We choose three environments that are challenging for all methods and cover different types of difficulty: \texttt{antmaze-giant} for long-horizon navigation, \texttt{cube-single} and \texttt{puzzle-4x4} for manipulation tasks. Across all three, the performance of \ours{} suggests that the predictive representation helps it learn useful features from fewer samples. On \texttt{puzzle-4x4}, \ours{} keeps most of its performance down to half of the data, while every baseline stays low at every fraction. Among the baselines, HIQL stands out on \texttt{antmaze-giant}, and GCIQL performs similarly to \ours{} on \texttt{cube-single}.

\paragraph{Failure Analysis (RQ4).}\label{sec:exp_failure} We observe two situations where \ours{} is less effective. The first is visual bias in pixel-based navigation: in \texttt{visual-antmaze}, the agent sometimes reaches a state that looks almost identical to the goal but lies in a different part of the maze, and the critic assigns it a high value (Appendix~\ref{sec:failure-case}, Figure~\ref{fig:valn_fail}); in \texttt{puzzle-3x3}, the policy can also settle into a repeated action loop (Figure~\ref{fig:puzzle_fail}). The second arises in state-based tasks such as \texttt{cube-double} and \texttt{scene}. There, methods that restrict learning to dataset actions, such as GCIQL, GCIVL, and $\text{Dual}_\text{GCIVL}$, are better suited, since they learn Q-values only from actions in the dataset, which limits how far errors can spread~\citep{kostrikov2021offlinereinforcementlearningimplicit, kumar2020conservative}. The GCTD3 backbone shows the same failure pattern as \ours{} (Table~\ref{tab:td3}), suggesting that some of the limitations of the backbone carry over to \ours{} and are not fully addressed by the learned representation.
\cutsectionup
\section{Conclusion}
\label{sec:conclusion}
\cutsectionup
From a representation learning perspective, offline GCRL demands representations that captures the environment at more than one scale, not just effective policies or well tuned value functions. \ours{} learns such a representation by predicting the information that matters for reaching the goal: how the state dynamics changes, how the return changes toward the goal, and which actions lead to it. Across state-based and pixel-based tasks, this representation gives a clear improvement over its own backbone and over prior goal-representation methods. The alignment conditions improve performance, specifically temporal alignment matters most for long-horizon navigation, behavioral alignment matters most for pixel-based tasks, and dynamical alignment gives a smaller gain spread across most tasks. This supports learning the representation from several predictive signals at once.

\paragraph{Limitations.} On some long-horizon tasks the representation alone is not enough and explicit planning helps, and on tasks such as \texttt{cube-double} and \texttt{scene}, methods that stay close to the dataset perform better. Since \ours{} does not depend on the policy that uses its representation, combining it with hierarchical planning is a natural next step.


\bibliography{iclr2027_conference}

@misc{alemi2019vib,
      title={Deep Variational Information Bottleneck}, 
      author={Alexander A. Alemi and Ian Fischer and Joshua V. Dillon and Kevin Murphy},
      year={2019},
      eprint={1612.00410},
      archivePrefix={arXiv},
      primaryClass={cs.LG},
      url={https://arxiv.org/abs/1612.00410}, 
}

@misc{fujimoto2025mrq,
      title={Towards General-Purpose Model-Free Reinforcement Learning}, 
      author={Scott Fujimoto and Pierluca D'Oro and Amy Zhang and Yuandong Tian and Michael Rabbat},
      year={2025},
      eprint={2501.16142},
      archivePrefix={arXiv},
      primaryClass={cs.LG},
      url={https://arxiv.org/abs/2501.16142}, 
}

@inproceedings{fujimoto2021minimalist,
	title={A Minimalist Approach to Offline Reinforcement Learning},
	author={Scott Fujimoto and Shixiang Shane Gu},
	booktitle={Thirty-Fifth Conference on Neural Information Processing Systems},
	year={2021},
}

@inproceedings{fujimoto2018addressing,
  title={Addressing function approximation error in actor-critic methods},
  author={Fujimoto, Scott and Hoof, Herke and Meger, David},
  booktitle={International conference on machine learning},
  pages={1587--1596},
  year={2018},
  organization={PMLR}
}

@misc{kostrikov2021offlinereinforcementlearningimplicit,
      title={Offline Reinforcement Learning with Implicit Q-Learning}, 
      author={Ilya Kostrikov and Ashvin Nair and Sergey Levine},
      year={2021},
      eprint={2110.06169},
      archivePrefix={arXiv},
      primaryClass={cs.LG},
      url={https://arxiv.org/abs/2110.06169}, 
}

@misc{park2024foundationpolicieshilbertrepresentations,
      title={Foundation Policies with Hilbert Representations}, 
      author={Seohong Park and Tobias Kreiman and Sergey Levine},
      year={2024},
      eprint={2402.15567},
      archivePrefix={arXiv},
      primaryClass={cs.LG},
      url={https://arxiv.org/abs/2402.15567}, 
}

@misc{park2024hiqlofflinegoalconditionedrl,
      title={HIQL: Offline Goal-Conditioned RL with Latent States as Actions}, 
      author={Seohong Park and Dibya Ghosh and Benjamin Eysenbach and Sergey Levine},
      year={2024},
      eprint={2307.11949},
      archivePrefix={arXiv},
      primaryClass={cs.LG},
      url={https://arxiv.org/abs/2307.11949}, 
}

@article{park2024ogbench,
  title={Ogbench: Benchmarking offline goal-conditioned rl},
  author={Park, Seohong and Frans, Kevin and Eysenbach, Benjamin and Levine, Sergey},
  journal={arXiv preprint arXiv:2410.20092},
  year={2024}
}

@article{kumar2020conservative,
  title={Conservative q-learning for offline reinforcement learning},
  author={Kumar, Aviral and Zhou, Aurick and Tucker, George and Levine, Sergey},
  journal={Advances in neural information processing systems},
  volume={33},
  pages={1179--1191},
  year={2020}
}

@article{hafner2022deep,
  title={Deep hierarchical planning from pixels},
  author={Hafner, Danijar and Lee, Kuang-Huei and Fischer, Ian and Abbeel, Pieter},
  journal={Advances in Neural Information Processing Systems},
  volume={35},
  pages={26091--26104},
  year={2022}
}

@misc{gcivl,
      title={Is Value Learning Really the Main Bottleneck in Offline RL?}, 
      author={Seohong Park and Kevin Frans and Sergey Levine and Aviral Kumar},
      year={2024},
      eprint={2406.09329},
      archivePrefix={arXiv},
      primaryClass={cs.LG},
      url={https://arxiv.org/abs/2406.09329}, 
}

@misc{gcbc1,
      title={Learning Latent Plans from Play}, 
      author={Corey Lynch and Mohi Khansari and Ted Xiao and Vikash Kumar and Jonathan Tompson and Sergey Levine and Pierre Sermanet},
      year={2019},
      eprint={1903.01973},
      archivePrefix={arXiv},
      primaryClass={cs.RO},
      url={https://arxiv.org/abs/1903.01973}, 
}

@misc{crl,
      title={Contrastive Learning as Goal-Conditioned Reinforcement Learning}, 
      author={Benjamin Eysenbach and Tianjun Zhang and Ruslan Salakhutdinov and Sergey Levine},
      year={2023},
      eprint={2206.07568},
      archivePrefix={arXiv},
      primaryClass={cs.LG},
      url={https://arxiv.org/abs/2206.07568}, 
}

@article{tdmpc2,
  title={Td-mpc2: Scalable, robust world models for continuous control},
  author={Hansen, Nicklas and Su, Hao and Wang, Xiaolong},
  journal={arXiv preprint arXiv:2310.16828},
  year={2023}
}

@article{dreamerv3,
  title={Mastering diverse domains through world models},
  author={Hafner, Danijar and Pasukonis, Jurgis and Ba, Jimmy and Lillicrap, Timothy},
  journal={arXiv preprint arXiv:2301.04104},
  year={2023}
}

@misc{qrl,
      title={Optimal Goal-Reaching Reinforcement Learning via Quasimetric Learning}, 
      author={Tongzhou Wang and Antonio Torralba and Phillip Isola and Amy Zhang},
      year={2023},
      eprint={2304.01203},
      archivePrefix={arXiv},
      primaryClass={cs.LG},
      url={https://arxiv.org/abs/2304.01203}, 
}

@inproceedings{Kaelbling1993LearningTA,
  title={Learning to Achieve Goals},
  author={Leslie Pack Kaelbling},
  booktitle={International Joint Conference on Artificial Intelligence},
  year={1993},
  url={https://api.semanticscholar.org/CorpusID:5538688}
}

@misc{yang2022rethinkinggoalconditionedsupervisedlearning,
      title={Rethinking Goal-conditioned Supervised Learning and Its Connection to Offline RL}, 
      author={Rui Yang and Yiming Lu and Wenzhe Li and Hao Sun and Meng Fang and Yali Du and Xiu Li and Lei Han and Chongjie Zhang},
      year={2022},
      eprint={2202.04478},
      archivePrefix={arXiv},
      primaryClass={cs.LG},
      url={https://arxiv.org/abs/2202.04478}, 
}

@misc{schwarzer2021,
      title={Data-Efficient Reinforcement Learning with Self-Predictive Representations}, 
      author={Max Schwarzer and Ankesh Anand and Rishab Goel and R Devon Hjelm and Aaron Courville and Philip Bachman},
      year={2021},
      eprint={2007.05929},
      archivePrefix={arXiv},
      primaryClass={cs.LG},
      url={https://arxiv.org/abs/2007.05929}, 
}

@misc{srinivas2020curlcontrastiveunsupervisedrepresentations,
      title={CURL: Contrastive Unsupervised Representations for Reinforcement Learning}, 
      author={Aravind Srinivas and Michael Laskin and Pieter Abbeel},
      year={2020},
      eprint={2004.04136},
      archivePrefix={arXiv},
      primaryClass={cs.LG},
      url={https://arxiv.org/abs/2004.04136}, 
}

@misc{gelu,
      title={Gaussian Error Linear Units (GELUs)}, 
      author={Dan Hendrycks and Kevin Gimpel},
      year={2023},
      eprint={1606.08415},
      archivePrefix={arXiv},
      primaryClass={cs.LG},
      url={https://arxiv.org/abs/1606.08415}, 
}

@misc{dual_goal,
      title={Dual Goal Representations}, 
      author={Seohong Park and Deepinder Mann and Sergey Levine},
      year={2025},
      eprint={2510.06714},
      archivePrefix={arXiv},
      primaryClass={cs.LG},
      url={https://arxiv.org/abs/2510.06714}, 
}

@misc{TRA,
      title={Temporal Representation Alignment: Successor Features Enable Emergent Compositionality in Robot Instruction Following}, 
      author={Vivek Myers and Bill Chunyuan Zheng and Anca Dragan and Kuan Fang and Sergey Levine},
      year={2025},
      eprint={2502.05454},
      archivePrefix={arXiv},
      primaryClass={cs.RO},
      url={https://arxiv.org/abs/2502.05454}, 
}

@misc{byol,
      title={Self-Predictive Representations for Combinatorial Generalization in Behavioral Cloning}, 
      author={Daniel Lawson and Adriana Hugessen and Charlotte Cloutier and Glen Berseth and Khimya Khetarpal},
      year={2025},
      eprint={2506.10137},
      archivePrefix={arXiv},
      primaryClass={cs.LG},
      url={https://arxiv.org/abs/2506.10137}, 
}

@inproceedings{litman2001,
author = {Littman, Michael L. and Sutton, Richard S. and Singh, Satinder},
title = {Predictive representations of state},
year = {2001},
publisher = {MIT Press},
address = {Cambridge, MA, USA},
booktitle = {Proceedings of the 15th International Conference on Neural Information Processing Systems: Natural and Synthetic},
pages = {1555–1561},
numpages = {7},
location = {Vancouver, British Columbia, Canada},
series = {NIPS'01}
}

@inproceedings{parr2008,
author = {Parr, Ronald and Li, Lihong and Taylor, Gavin and Painter-Wakefield, Christopher and Littman, Michael L.},
title = {An analysis of linear models, linear value-function approximation, and feature selection for reinforcement learning},
year = {2008},
isbn = {9781605582054},
publisher = {Association for Computing Machinery},
address = {New York, NY, USA},
url = {https://doi.org/10.1145/1390156.1390251},
doi = {10.1145/1390156.1390251},
booktitle = {Proceedings of the 25th International Conference on Machine Learning},
pages = {752–759},
numpages = {8},
location = {Helsinki, Finland},
series = {ICML '08}
}

@misc{gelada2019deepmdplearningcontinuouslatent,
      title={DeepMDP: Learning Continuous Latent Space Models for Representation Learning}, 
      author={Carles Gelada and Saurabh Kumar and Jacob Buckman and Ofir Nachum and Marc G. Bellemare},
      year={2019},
      eprint={1906.02736},
      archivePrefix={arXiv},
      primaryClass={cs.LG},
      url={https://arxiv.org/abs/1906.02736}, 
}

@article{ha2018world,
  title={World models},
  author={Ha, David and Schmidhuber, J{\"u}rgen},
  journal={arXiv preprint arXiv:1803.10122},
  volume={2},
  number={3},
  pages={440},
  year={2018}
}

@misc{Dreamerv1,
      title={Dream to Control: Learning Behaviors by Latent Imagination}, 
      author={Danijar Hafner and Timothy Lillicrap and Jimmy Ba and Mohammad Norouzi},
      year={2020},
      eprint={1912.01603},
      archivePrefix={arXiv},
      primaryClass={cs.LG},
      url={https://arxiv.org/abs/1912.01603}, 
}

@article{Schrittwieser_2020,
   title={Mastering Atari, Go, chess and shogi by planning with a learned model},
   volume={588},
   ISSN={1476-4687},
   url={http://dx.doi.org/10.1038/s41586-020-03051-4},
   DOI={10.1038/s41586-020-03051-4},
   number={7839},
   journal={Nature},
   publisher={Springer Science and Business Media LLC},
   author={Schrittwieser, Julian and Antonoglou, Ioannis and Hubert, Thomas and Simonyan, Karen and Sifre, Laurent and Schmitt, Simon and Guez, Arthur and Lockhart, Edward and Hassabis, Demis and Graepel, Thore and Lillicrap, Timothy and Silver, David},
   year={2020},
   month=dec, pages={604–609} }

@misc{planet,
      title={Learning Latent Dynamics for Planning from Pixels}, 
      author={Danijar Hafner and Timothy Lillicrap and Ian Fischer and Ruben Villegas and David Ha and Honglak Lee and James Davidson},
      year={2019},
      eprint={1811.04551},
      archivePrefix={arXiv},
      primaryClass={cs.LG},
      url={https://arxiv.org/abs/1811.04551}, 
}

@misc{vip,
      title={VIP: Towards Universal Visual Reward and Representation via Value-Implicit Pre-Training}, 
      author={Yecheng Jason Ma and Shagun Sodhani and Dinesh Jayaraman and Osbert Bastani and Vikash Kumar and Amy Zhang},
      year={2023},
      eprint={2210.00030},
      archivePrefix={arXiv},
      primaryClass={cs.RO},
      url={https://arxiv.org/abs/2210.00030}, 
}

@misc{adamw,
      title={Decoupled Weight Decay Regularization}, 
      author={Ilya Loshchilov and Frank Hutter},
      year={2019},
      eprint={1711.05101},
      archivePrefix={arXiv},
      primaryClass={cs.LG},
      url={https://arxiv.org/abs/1711.05101}, 
}

@misc{impala,
      title={IMPALA: Scalable Distributed Deep-RL with Importance Weighted Actor-Learner Architectures}, 
      author={Lasse Espeholt and Hubert Soyer and Remi Munos and Karen Simonyan and Volodymir Mnih and Tom Ward and Yotam Doron and Vlad Firoiu and Tim Harley and Iain Dunning and Shane Legg and Koray Kavukcuoglu},
      year={2018},
      eprint={1802.01561},
      archivePrefix={arXiv},
      primaryClass={cs.LG},
      url={https://arxiv.org/abs/1802.01561}, 
}

@misc{park_long_horizon,
      title={Horizon Reduction Makes RL Scalable}, 
      author={Seohong Park and Kevin Frans and Deepinder Mann and Benjamin Eysenbach and Aviral Kumar and Sergey Levine},
      year={2025},
      eprint={2506.04168},
      archivePrefix={arXiv},
      primaryClass={cs.LG},
      url={https://arxiv.org/abs/2506.04168}, 
}

@misc{paszke2019pytorchimperativestylehighperformance,
      title={PyTorch: An Imperative Style, High-Performance Deep Learning Library}, 
      author={Adam Paszke and Sam Gross and Francisco Massa and Adam Lerer and James Bradbury and Gregory Chanan and Trevor Killeen and Zeming Lin and Natalia Gimelshein and Luca Antiga and Alban Desmaison and Andreas Köpf and Edward Yang and Zach DeVito and Martin Raison and Alykhan Tejani and Sasank Chilamkurthy and Benoit Steiner and Lu Fang and Junjie Bai and Soumith Chintala},
      year={2019},
      eprint={1912.01703},
      archivePrefix={arXiv},
      primaryClass={cs.LG},
      url={https://arxiv.org/abs/1912.01703}, 
}

@inproceedings{kakade2002approximately,
  title     = {Approximately Optimal Approximate Reinforcement Learning},
  author    = {Kakade, Sham and Langford, John},
  booktitle = {Proceedings of the 19th International Conference on Machine Learning (ICML)},
  pages     = {267--274},
  year      = {2002}
}

@inproceedings{ross2011reduction,
  title     = {A Reduction of Imitation Learning and Structured Prediction to No-Regret Online Learning},
  author    = {Ross, St{\'e}phane and Gordon, Geoffrey and Bagnell, Drew},
  booktitle = {Proceedings of the 14th International Conference on Artificial Intelligence and Statistics (AISTATS)},
  pages     = {627--635},
  year      = {2011}
}

@article{Sutton:1988,
  author    = {Richard S. Sutton},
  title     = {Learning to Predict by the Methods of Temporal Differences},
  journal   = {Machine Learning},
  volume    = {3},
  number    = {1},
  pages     = {9--44},
  year      = {1988},
  month     = {August},
  publisher = {Springer},
  doi       = {10.1007/BF00115009}
}

@misc{quillen2018,
      title={Deep Reinforcement Learning for Vision-Based Robotic Grasping: A Simulated Comparative Evaluation of Off-Policy Methods}, 
      author={Deirdre Quillen and Eric Jang and Ofir Nachum and Chelsea Finn and Julian Ibarz and Sergey Levine},
      year={2018},
      eprint={1802.10264},
      archivePrefix={arXiv},
      primaryClass={cs.RO},
      url={https://arxiv.org/abs/1802.10264}, 
}

@misc{liu2022gcrlproblems,
      title={Goal-Conditioned Reinforcement Learning: Problems and Solutions}, 
      author={Minghuan Liu and Menghui Zhu and Weinan Zhang},
      year={2022},
      eprint={2201.08299},
      archivePrefix={arXiv},
      primaryClass={cs.AI},
      url={https://arxiv.org/abs/2201.08299}, 
}

@misc{bardes2024revisitingfeaturepredictionlearning,
      title={Revisiting Feature Prediction for Learning Visual Representations from Video}, 
      author={Adrien Bardes and Quentin Garrido and Jean Ponce and Xinlei Chen and Michael Rabbat and Yann LeCun and Mahmoud Assran and Nicolas Ballas},
      year={2024},
      eprint={2404.08471},
      archivePrefix={arXiv},
      primaryClass={cs.CV},
      url={https://arxiv.org/abs/2404.08471}, 
}

@article{li2006towards,
  title={Towards a unified theory of state abstraction for MDPs.},
  author={Li, Lihong and Walsh, Thomas J and Littman, Michael L},
  journal={AI\&M},
  volume={1},
  number={2},
  pages={3},
  year={2006}
}

@misc{castro2019scalablemethodscomputingstate,
      title={Scalable methods for computing state similarity in deterministic Markov Decision Processes}, 
      author={Pablo Samuel Castro},
      year={2019},
      eprint={1911.09291},
      archivePrefix={arXiv},
      primaryClass={cs.LG},
      url={https://arxiv.org/abs/1911.09291}, 
}

@misc{lei2025qwsloptimizinggoalconditionedrl,
      title={Q-WSL: Optimizing Goal-Conditioned RL with Weighted Supervised Learning via Dynamic Programming}, 
      author={Xing Lei and Xuetao Zhang and Zifeng Zhuang and Donglin Wang},
      year={2025},
      eprint={2410.06648},
      archivePrefix={arXiv},
      primaryClass={cs.LG},
      url={https://arxiv.org/abs/2410.06648}, 
}
\bibliographystyle{iclr2027_conference}

\appendix
\clearpage
\section{Proofs and Extended Analysis}
\label{app:proofs}
Notation is that of Sections~\ref{sec:preliminary}--\ref{sec:method}.
All expectations are over $(s, a, s', g) \sim \mathcal{D}$ and, wherever $r_{\mathrm{MC}}$ appears, over the remainder of the dataset trajectory that determines it.
The latents $\mathbf{z}_s = \mathcal{E}^s_\psi(s)$, $\mathbf{z}_g = \mathcal{E}^s_\psi(g)$, and $\mathbf{z}_{sa} = \mathcal{E}^{sa}_\psi(\mathbf{z}_s, a)$ are deterministic functions of $(s, a, g)$, and $\mathbf{z}_{s'} = \mathcal{E}^s_{\bar\psi}(s')$ is a deterministic function of $s'$.
We write $\|\cdot\|$ for the Euclidean norm and $\|Y\|_{L^2} := (\E[Y^2])^{1/2}$ for a scalar random variable $Y$.
Rewards are bounded and $\gamma < 1$, so $r_{\mathrm{MC}}$ and all value functions are square-integrable.

\paragraph{Two standard facts.}
We use these repeatedly; both are textbook properties of conditional expectation.
\begin{enumerate}
  \item[\textbf{F1}] \emph{Conditional mean is the $L^2$ projection.}
  For a square-integrable random variable $Y$ and a random vector $W$,
  $\min_{f} \E[(Y - f(W))^2]$ over measurable $f$ is attained at $f(W) = \E[Y \mid W]$.
  \item[\textbf{F2}] \emph{Tower property.}
  If $W$ is a function of $X$, then $\E\big[\E[Y \mid X] \,\big|\, W\big] = \E[Y \mid W]$.
\end{enumerate}

\subsection{Proof of Proposition~\ref{prop:dyn}}
\label{app:proof_dyn}

\begin{proof}
Fix a sample $(s, a, s', g)$. Since $V$ is $L_V$-Lipschitz in its first argument,
\[
  \big|V(f_{\mathrm{dyn}}(\mathbf{z}_{sa}), \mathbf{z}_g) - V(\mathbf{z}_{s'}, \mathbf{z}_g)\big|
  \;\le\; L_V \,\big\|f_{\mathrm{dyn}}(\mathbf{z}_{sa}) - \mathbf{z}_{s'}\big\|.
\]
Squaring both sides and taking expectations,
\[
  \E\Big[\big(V(f_{\mathrm{dyn}}(\mathbf{z}_{sa}), \mathbf{z}_g) - V(\mathbf{z}_{s'}, \mathbf{z}_g)\big)^2\Big]
  \;\le\; L_V^2\, \E\big[\|f_{\mathrm{dyn}}(\mathbf{z}_{sa}) - \mathbf{z}_{s'}\|^2\big]
  \;=\; L_V^2\, \delta_{\mathrm{dyn}},
\]
where the last equality is the definition \eqref{eq:delta_dyn}.
\end{proof}

\subsection{Proof of Proposition~\ref{prop:temp}}
\label{app:proof_temp}

Throughout this proof write
\[
  X := (s, a, g), \qquad W := (\mathbf{z}_{sa}, \mathbf{z}_g), \qquad \mu(W) := \E[r_{\mathrm{MC}} \mid W].
\]
$W$ is a function of $X$, and by definition $Q^\beta(X) = \E[r_{\mathrm{MC}} \mid X]$.

\begin{lemma}
\label{lem:mu}
$\mu(W) = \E[Q^\beta \mid W]$.
\end{lemma}
\begin{proof}
By (F2) applied with $Y = r_{\mathrm{MC}}$,
$\E[r_{\mathrm{MC}} \mid W] = \E\big[\E[r_{\mathrm{MC}} \mid X] \,\big|\, W\big] = \E[Q^\beta \mid W]$.
\end{proof}

\begin{proof}[Proof of Proposition~\ref{prop:temp}]
We establish the identity $\Delta_{\mathrm{temp}} = \sigma^2_{\mathrm{MC}} + \varepsilon_V^{\min}$ in four steps.\\

\smallskip\noindent
\textbf{Step 1: $\Delta_{\mathrm{temp}} = \E[(r_{\mathrm{MC}} - \mu(W))^2]$.}
Definition \eqref{eq:delta_temp} minimizes $\E[(r_{\mathrm{MC}} - f(W))^2]$ over measurable $f$. By (F1) with $Y = r_{\mathrm{MC}}$, the minimizer is $f(W) = \E[r_{\mathrm{MC}} \mid W] = \mu(W)$.\\

\smallskip\noindent
\textbf{Step 2: $\varepsilon_V^{\min} = \E[(Q^\beta - \mu(W))^2]$.}
The definition of $\varepsilon_V^{\min}$ (Sec-\ref{sec:temp_align}) minimizes $\E[(Q^\beta - Q(W))^2]$ over measurable $Q$. By (F1) with $Y = Q^\beta$, the minimizer is $Q(W) = \E[Q^\beta \mid W]$, which equals $\mu(W)$ by Lemma~\ref{lem:mu}.\\

\smallskip\noindent
\textbf{Step 3: the cross term vanishes,}
$\E\big[(r_{\mathrm{MC}} - Q^\beta)(Q^\beta - \mu(W))\big] = 0$.
The factor $Q^\beta - \mu(W)$ is a function of $X$ (both $Q^\beta$ and $W$ are). Conditioning on $X$ and using (F2),
\[
  \E\big[(r_{\mathrm{MC}} - Q^\beta)(Q^\beta - \mu(W))\big]
  = \E\Big[(Q^\beta - \mu(W))\cdot \underbrace{\E[r_{\mathrm{MC}} - Q^\beta \mid X]}_{=\,Q^\beta - Q^\beta\,=\,0}\Big]
  = 0.
\]

\smallskip\noindent
\textbf{Step 4: assemble.}
Write $r_{\mathrm{MC}} - \mu(W) = (r_{\mathrm{MC}} - Q^\beta) + (Q^\beta - \mu(W))$, square, and take expectations. The cross term is zero by Step 3, so
\[
  \underbrace{\E[(r_{\mathrm{MC}} - \mu(W))^2]}_{\Delta_{\mathrm{temp}}\ \text{(Step 1)}}
  \;=\;
  \underbrace{\E[(r_{\mathrm{MC}} - Q^\beta)^2]}_{\sigma^2_{\mathrm{MC}}}
  \;+\;
  \underbrace{\E[(Q^\beta - \mu(W))^2]}_{\varepsilon_V^{\min}\ \text{(Step 2)}} .
\]
This is \eqref{eq:prop_temp}.
\end{proof}

\paragraph{Why the trained loss is an upper bound.}
A specific trained head $f_{\mathrm{rew}}$ is one particular measurable function of $W$, so $\E[\mathcal{L}_{\mathrm{rew}}] = \E[(r_{\mathrm{MC}} - f_{\mathrm{rew}}(W))^2] \ge \Delta_{\mathrm{temp}}$, with equality only if $f_{\mathrm{rew}} = \mu$ almost surely. This is the inequality in \eqref{eq:delta_temp}.

\subsection{Representation error versus data error}
\label{app:proof_cor}

\begin{corollary}[Representation error versus data error]
\label{cor:data_gap}
Let $\varepsilon_\beta := \big(\E[(Q^\beta - Q^*)^2]\big)^{1/2}$. Then
\begin{equation}
  \min_{Q}\ \Big(\E\big[\big(Q(\mathbf{z}_{sa}, \mathbf{z}_g) - Q^*(s,a,g)\big)^2\big]\Big)^{1/2}
  \;\le\; \sqrt{\varepsilon_V^{\min}} \;+\; \varepsilon_\beta.
  \label{eq:cor_data_gap}
\end{equation}
\end{corollary}
\noindent
The error against the \emph{optimal} value splits into a representation term, controlled by the encoder through Proposition~\ref{prop:temp}, and a data term $\varepsilon_\beta$, the behavior policy's suboptimality, which no encoder can reduce.
As the behavior data degrades, for instance on stitching datasets where individual behavior trajectories do not reach the goal, $\varepsilon_\beta$ grows while the representation term does not, which is the mechanism behind the graceful, rather than abrupt, degradation reported in Sec-\ref{sec:exp_robustness}.

\begin{proof}
For any critic $Q$ on $W$, Minkowski's inequality in $L^2$ gives
\[
  \|Q(W) - Q^*\|_{L^2}
  \;\le\; \|Q(W) - Q^\beta\|_{L^2} + \|Q^\beta - Q^*\|_{L^2}
  \;=\; \|Q(W) - Q^\beta\|_{L^2} + \varepsilon_\beta .
\]
The second term does not depend on $Q$. Taking the minimum over $Q$ on both sides and using $\min_Q \|Q(W) - Q^\beta\|_{L^2} = \sqrt{\varepsilon_V^{\min}}$ (the square root of the definition of $\varepsilon_V^{\min}$ in Sec-\ref{sec:temp_align}) yields \eqref{eq:cor_data_gap}.
\end{proof}

\subsection{When the condition binds}
\label{app:proof_scope}

\begin{remark}[When the condition binds]
\label{rem:scope}
If $\mathcal{E}^s_\psi$ and $\mathcal{E}^{sa}_\psi$ are injective on the data support, as when raw states are used with no compression, then $(\mathbf{z}_{sa}, \mathbf{z}_g)$ determines $(s,a,g)$ and $\varepsilon_V^{\min} = 0$ with no auxiliary objective at all.
Propositions~\ref{prop:dyn}--\ref{prop:beh} are therefore statements about \emph{learned, compressive} encoders: the regime of pixel observations and high-dimensional state.
\end{remark}

\begin{proof}
If $\mathcal{E}^s_\psi$ and $\mathcal{E}^{sa}_\psi$ are injective on the data support, the map $X \mapsto W$ is injective there, so $W$ determines $X$ and the two generate the same information: $\E[Q^\beta \mid W] = Q^\beta$. By Step~2 of the proof of Proposition~\ref{prop:temp}, $\varepsilon_V^{\min} = \E[(Q^\beta - \E[Q^\beta \mid W])^2] = 0$.
\end{proof}

\subsection{Proof of Proposition~\ref{prop:beh}}
\label{app:proof_beh}

\begin{proof}
Fix $(s, a, g)$ and write $a^* := \pi^*(s, g)$ and $\hat{a} := f_{\mathrm{g\text{-}act}}(\mathbf{z}_s, \mathbf{z}_g)$.
Because $a^*$ maximizes $Q^*(s, \cdot, g)$, the gap $Q^*(s, a^*, g) - Q^*(s, \hat{a}, g)$ is nonnegative, and because $Q^*$ is $L_Q$-Lipschitz in its action argument,
\[
  0 \;\le\; Q^*(s, a^*, g) - Q^*(s, \hat{a}, g) \;\le\; L_Q\, \|a^* - \hat{a}\|.
\]
Insert the dataset action $a$ and apply the triangle inequality:
\[
  \|a^* - \hat{a}\| \;\le\; \|a^* - a\| + \|a - \hat{a}\|.
\]
Take expectations over $(s, a, g) \sim \mathcal{D}$. For each term, Jensen's inequality gives $\E\|Y\| \le (\E\|Y\|^2)^{1/2}$, hence
\[
  \E\big[Q^*(s, a^*, g) - Q^*(s, \hat{a}, g)\big]
  \;\le\; L_Q\Big(\E\|a^* - a\| + \E\|a - \hat{a}\|\Big)
  \;\le\; L_Q\Big(\sqrt{\varepsilon_\pi} + \sqrt{\delta_{\mathrm{beh}}}\Big),
\]
where $\E\|a - \hat{a}\|^2 = \delta_{\mathrm{beh}}$ by \eqref{eq:delta_beh} and $\E\|a^* - a\|^2 = \varepsilon_\pi$ by definition. This is \eqref{eq:prop_beh}.
\end{proof}

\paragraph{Multi-step form (standard; stated for completeness, not used in the main text).}
Let $\epsilon := \E[Q^*(s, \pi^*(s,g), g) - Q^*(s, \hat{a}, g)]$ denote the one-step gap bounded above.
If the same gap holds under the state distribution induced by the latent policy $\hat{a}$, the performance-difference lemma~\citep{kakade2002approximately} gives a return gap of at most $\epsilon/(1-\gamma)$.
When the gap is only controlled under the dataset distribution $\mathcal{D}$, as here, the additional distribution shift between $\mathcal{D}$ and the latent policy's occupancy contributes a further factor of $1/(1-\gamma)$, giving $O(\epsilon/(1-\gamma)^2)$~\citep{ross2011reduction}.
Both factors are properties of the compounding-error argument, not of the representation; Proposition~\ref{prop:beh} controls only $\epsilon$.

\subsection{Goal-level dynamics introduces no fourth condition}
\label{app:proof_gdyn}

\begin{remark}[Goal-level dynamics]
\label{rem:gdyn}
$f_{\mathrm{g\text{-}dyn}}$ supervises the same latent pair as $f_{\mathrm{g\text{-}act}}$ with a complementary target, the successor latent.
Its achievable error is controlled by the other two conditions: if $a \mapsto f_{\mathrm{dyn}}(\mathcal{E}^{sa}_\psi(\mathbf{z}_s, a))$ is $L_{\mathrm{dyn}}$-Lipschitz, then
\begin{equation}
\min_f \E\|f(\mathbf{z}_s, \mathbf{z}_g) - \mathbf{z}_{s'}\|_2^2 \;\le\; 2\,\delta_{\mathrm{dyn}} + 2\,L_{\mathrm{dyn}}^2\, \delta_{\mathrm{beh}}.
\end{equation}
$f_{\mathrm{g\text{-}dyn}}$ therefore introduces no fourth condition; it supervises directly a quantity the other heads reach only by composition. Direct supervision matters in practice: removing it is the largest single loss on pixel-based tasks (Table~\ref{tab:ablation}, Sec-\ref{sec:ablations}).
\end{remark}

\begin{proof}
Define $h_s(a) := f_{\mathrm{dyn}}\big(\mathcal{E}^{sa}_\psi(\mathbf{z}_s, a)\big)$, the forward-dynamics prediction as a function of the action for a fixed $\mathbf{z}_s$; by assumption $h_s$ is $L_{\mathrm{dyn}}$-Lipschitz. Note that for the dataset action, $h_s(a) = f_{\mathrm{dyn}}(\mathbf{z}_{sa})$.
Consider the candidate goal-level dynamics head
\[
  \bar{f}(\mathbf{z}_s, \mathbf{z}_g) \;:=\; h_s\big(\hat{a}\big), \qquad \hat{a} = f_{\mathrm{g\text{-}act}}(\mathbf{z}_s, \mathbf{z}_g),
\]
which depends on $(\mathbf{z}_s, \mathbf{z}_g)$ only. By the triangle inequality and the Lipschitz property,
\[
  \|\bar{f}(\mathbf{z}_s, \mathbf{z}_g) - \mathbf{z}_{s'}\|
  \;\le\; \|h_s(\hat{a}) - h_s(a)\| + \|h_s(a) - \mathbf{z}_{s'}\|
  \;\le\; L_{\mathrm{dyn}}\,\|\hat{a} - a\| + \|f_{\mathrm{dyn}}(\mathbf{z}_{sa}) - \mathbf{z}_{s'}\| .
\]
Using $(x + y)^2 \le 2x^2 + 2y^2$ and taking expectations,
\[
  \E\|\bar{f}(\mathbf{z}_s, \mathbf{z}_g) - \mathbf{z}_{s'}\|^2
  \;\le\; 2\,L_{\mathrm{dyn}}^2\,\E\|\hat{a} - a\|^2 + 2\,\E\|f_{\mathrm{dyn}}(\mathbf{z}_{sa}) - \mathbf{z}_{s'}\|^2
  \;=\; 2\,L_{\mathrm{dyn}}^2\,\delta_{\mathrm{beh}} + 2\,\delta_{\mathrm{dyn}} .
\]
The minimum over all heads $f(\mathbf{z}_s, \mathbf{z}_g)$ is at most the error of this particular $\bar{f}$, which gives the stated bound.
\end{proof}

\subsection{The three terms are distinct}
\label{app:distinct}
None of $\delta_{\mathrm{dyn}}$, $\Delta_{\mathrm{temp}}$, $\delta_{\mathrm{beh}}$ controls another.
A latent can be transition-consistent yet discard goal distance encoding local pose but not maze position giving small $\delta_{\mathrm{dyn}}$ and large $\Delta_{\mathrm{temp}}$.
Return can be predictable from $(\mathbf{z}_{sa}, \mathbf{z}_g)$, which sees the action, while $(\mathbf{z}_s, \mathbf{z}_g)$ does not determine which action is goal-directed small $\Delta_{\mathrm{temp}}$, large $\delta_{\mathrm{beh}}$.
And knowing the goal-directed action does not make its consequence predictable small $\delta_{\mathrm{beh}}$, large $\delta_{\mathrm{dyn}}$.
Removing any one head leaves negatively impacts the overall performance Table~\ref{tab:ablation}.

\clearpage
\section{Environment Details}

Our main benchmark uses 17 environments from the OGBench~\citep{park2024ogbench} suite (12 state-based and 5 pixel-based), and we report 5 additional \texttt{stitch} datasets (3 state-based and 2 pixel-based) in Appendix~\ref{app:stitch}. 

\textbf{State-Based Environments (12 tasks):}
\begin{itemize}
    \item \textbf{Locomotion (7 tasks):}
    \begin{itemize}
        \item \texttt{pointmaze-medium-navigate}
        \item \texttt{pointmaze-large-navigate}
        \item \texttt{antmaze-medium-navigate}
        \item \texttt{antmaze-large-navigate}
        \item \texttt{antmaze-giant-navigate}
        \item \texttt{humanoidmaze-medium-navigate}
        \item \texttt{humanoidmaze-large-navigate}
    \end{itemize}
    \item \textbf{Manipulation (5 tasks):}
    \begin{itemize}
        \item \texttt{cube-single-play}
        \item \texttt{cube-double-play}
        \item \texttt{scene-play}
        \item \texttt{puzzle-3x3-play}
        \item \texttt{puzzle-4x4-play}
    \end{itemize}
\end{itemize}

\textbf{Pixel-Based Environments (5 tasks):}
\begin{itemize}
    \item \textbf{Locomotion (2 tasks):}
    \begin{itemize}
        \item \vamn
        \item \valn
    \end{itemize}
    \item \textbf{Manipulation (3 tasks):}
    \begin{itemize}
        \item \vcsp
        \item \vsp
        \item \vptp
    \end{itemize}
\end{itemize}

\textbf{Stitch Datasets (5 tasks, Appendix~\ref{app:stitch}):}
\begin{itemize}
    \item \textbf{State-based (3 tasks):} 
    \begin{itemize}
        \item \texttt{antmaze-medium-stitch}
        \item \texttt{humanoidmaze-medium-stitch}
        \item \texttt{pointmaze-medium-stitch}
    \end{itemize}
    \item \textbf{Pixel-based (2 tasks):} 
    \begin{itemize}
        \item \vams
        \item \vals
    \end{itemize}
\end{itemize}

\begin{figure}[h]
\centering
\begin{tikzpicture}
\node (a1) at (0,0) {\includegraphics[width=0.24\textwidth]{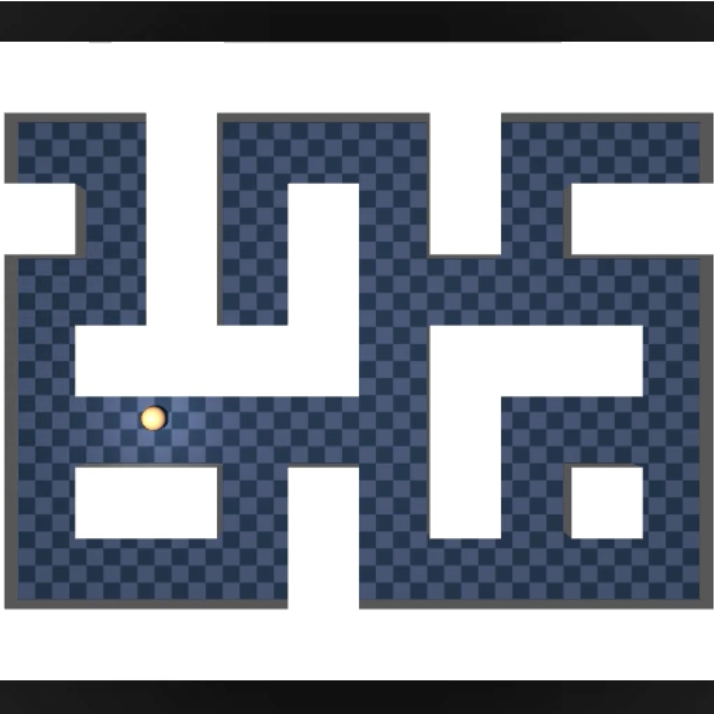}};
\node (b1) at (0.25\textwidth,0) {\includegraphics[width=0.24\textwidth]{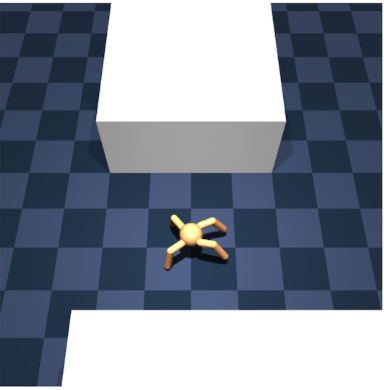}};
\node (c1) at (0.50\textwidth,0) {\includegraphics[width=0.24\textwidth]{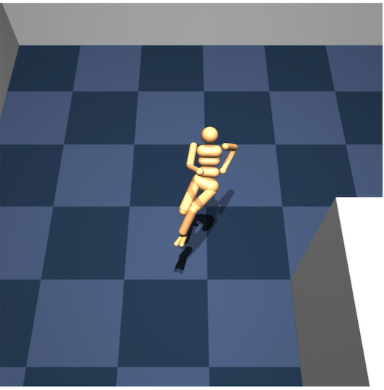}};
\node (d1) at (0.75\textwidth,0) {\includegraphics[width=0.24\textwidth]{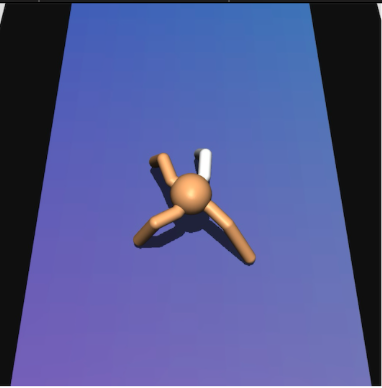}};
\node (a2) at (0,-0.25\textwidth) {\includegraphics[width=0.24\textwidth]{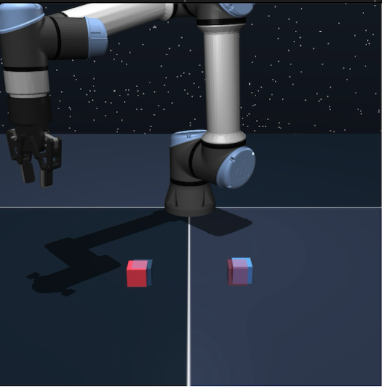}};
\node (b2) at (0.25\textwidth,-0.25\textwidth) {\includegraphics[width=0.24\textwidth]{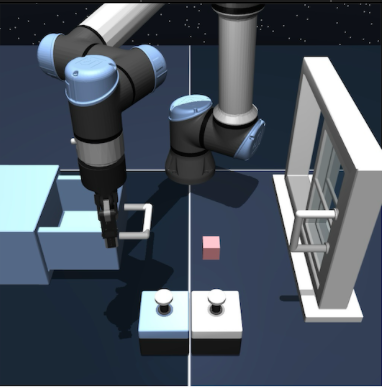}};
\node (c2) at (0.50\textwidth,-0.25\textwidth) {\includegraphics[width=0.24\textwidth]{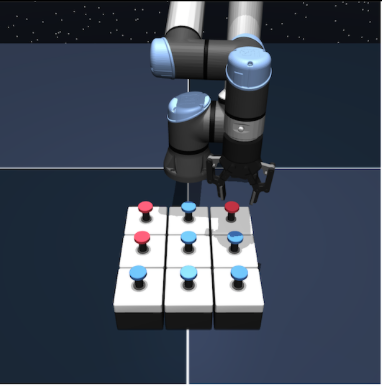}};
\node (d2) at (0.75\textwidth,-0.25\textwidth) {\includegraphics[width=0.24\textwidth]{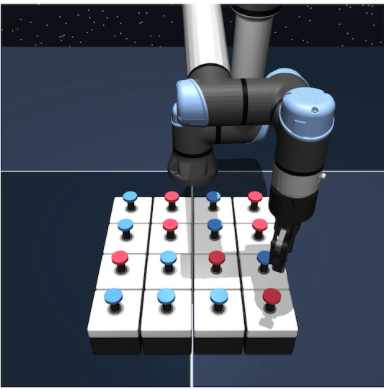}};
\end{tikzpicture}
\caption{\textbf{OGBench environments.} Navigation tasks include AntMaze, PointMaze, and HumanoidMaze; manipulation tasks include Cube, Scene, and Puzzle.}
\label{fig:envs}
\end{figure}

\clearpage
\section{Implementation Details} \label{appendix:exp_details}
\subsection{Hyperparameter}
\begin{table*}[ht]
\caption{
\footnotesize
\textbf{\ours{} Hyperparameters }
}
\label{table:hyp}
\footnotesize
\begin{tabular}{ll}
    \toprule
    \textbf{Hyperparameter} & \textbf{Value} \\
    \midrule
    \rowcolor{sb_orange!20}
    \multicolumn{2}{l}{\textbf{Agent}} \\
    \midrule
    Optimizer & AdamW~\citep{adamw} \\
    Learning rate & $3\text{e}{-4}$ \\
    Weight decay & $1\text{e}{-4}$ \\
    Gradient Steps & $1000000$ (states), $500000$ (pixels) \\
    Batch Size & $1024$ (states), $256$ (pixels) \\
    Policy MLP dimensions & $(512, 512, 512)$ \\
    Value MLP dimensions & $(512, 512, 512)$ \\
    Nonlinearity & GELU~\citep{gelu} \\
    Target update frequency $T_{\text{target}}$ & 250\\
    Target policy noise~$\sigma$      & $\mathcal{N}(0,0.2^2)$ \\
    Target policy noise clipping~$c$  & $(-0.1, 0.1)$ \\
    Discount factor $\gamma$ & $0.995$ (\texttt{antmaze-giant}, \texttt{humanoidmaze}), $0.99$ (others) \\
    Reward values & $\{-1, 0\}$ \\
    MSPR Prediction Horizon $H$ & 5\\
    Q-Horizon $n$ & 3\\
    Frame Stacking & $3$ (Pixel) \\
    BC-Coefficient $\lambda$ & $0.1$ (\texttt{Locomotion tasks}), $1.0$ (\texttt{Manipulation Tasks}) \\
    \midrule
    \rowcolor{sb_orange!20}
    \multicolumn{2}{l}{\textbf{Multi-stage Predictive Architecture}} \\
    \midrule
    Encoder module & MLP Block (state), IMPALA (pixel) \citep{impala}\\
    $\mathbf{z}_s$ dimension & 512\\
    $\mathbf{z}_{sa}$ dimension & 512 \\
    $\mathbf{z}_a$ dimension & 256 \\
    Hidden dimension & 512 \\
    Learning rate & $1\text{e}{-4}$ \\
    Weight decay & $1\text{e}{-4}$\\
    \midrule
    \rowcolor{sb_orange!20}
\multicolumn{2}{l}{\textbf{Multi-stage Predictors Loss}} \\
    \midrule
    Dynamics loss weight $\lambda_{\text{dyn}}$ & $1.0$ \\
    Return loss weight $\lambda_{\text{rew}}$ & $0.3$\\
    Goal-level dynamics loss weight $\lambda_{\text{g\text{-}dyn}}$ & $0.1$ \\
    Goal-conditioned action loss weight $\lambda_{\text{g\text{-}act}}$ & $0.1$  \\
    \bottomrule
\end{tabular}
\end{table*}

\clearpage
\subsection{Network Architecture}

This section describes the networks used in our method using PyTorch code \citep{paszke2019pytorchimperativestylehighperformance}. We utilize a standardized Multilayer Perceptron (MLP) block for vector inputs and an Impala CNN (Small)~\citep{impala, park2024ogbench} for image inputs. The state encoder, goal representation, value, and policy networks are described below. the backbone implementation is close to \citep{ fujimoto2021minimalist, fujimoto2025mrq}. 

\paragraph{Preamble}

We define a helper function to initialize MLPs. Unless specified otherwise, our MLPs utilize GELU activations. 

\begin{lstlisting}
import torch
import torch.nn as nn
import torch.nn.functional as F
from typing import Sequence

def apply_variance_scaling_init(net, scale=1.0):
    fan = (net.weight.size(-2) + net.weight.size(-1)) / 2
    init_w = math.sqrt(scale / fan)
    net.weight.data.uniform_(-init_w, init_w)
    net.bias.data.fill_(0)

def MLP(sizes, activation=nn.GELU, output_activation=nn.Identity, 
        layer_norm=True):
    layers = []
    for j in range(len(sizes) - 1):
        # Determine activation for current layer
        act = activation if j < len(sizes) - 2 else output_activation
        
        # Linear layer setup
        fc = nn.Linear(sizes[j], sizes[j + 1])
        
        # Apply LayerNorm (Pre-LN) to hidden layers (j > 0)
        if layer_norm and j > 0:
            layers += [nn.LayerNorm(sizes[j], eps=1e-6), fc, act()]
        else:
            layers += [fc, act()]
            
    return nn.Sequential(*layers)
\end{lstlisting}

\paragraph{Visual Encoder (Impala CNN)}

For image-based environments, we utilize the Impala CNN architecture similar to the one used in  OGBench Baselines ~\citep{park2024ogbench}. 

\begin{lstlisting}
class ResidualBlock(nn.Module):
    def __init__(self, in_ch, features):
        super().__init__()
        self.cnn1 = nn.Conv2d(in_ch, features, 3, stride=1, padding=1)
        self.cnn2 = nn.Conv2d(features, features, 3, stride=1, padding=1)
        self.cnn3 = nn.Conv2d(features, features, 3, stride=1, padding=1)
        self.max_pool = nn.MaxPool2d(3, stride=2, padding=1)

    def forward(self, x):
        x1 = self.max_pool(self.cnn1(x))
        x2 = self.cnn2(F.relu(x1))
        x2 = self.cnn3(F.relu(x2))
        return x1 + x2

class ImpalaCNN(nn.Module):
    def __init__(self, state_dim, out_dim):
        super().__init__()
        self.block1 = ResidualBlock(state_dim, 16)
        self.block2 = ResidualBlock(16, 32)
        self.block3 = ResidualBlock(32, 32)
        # Assumes standard OGBench input sizes 3x64x64, projecting to out_dim
        self.fc = MLP([2048, out_dim], output_activation=nn.GELU)
        
    def forward(self, x):
        x = x / 255.0 
        x = self.block1(x)
        x = self.block2(x)
        x = self.block3(x).reshape(x.shape[0], -1)
        return self.fc(x)
\end{lstlisting}

\paragraph{State/Goal Encoder}

For vector states, the state encoder $\mathcal{E}^s_\psi$ is a three-layer MLP with hidden dimensions of 512. For both state and goal the encoders are shared. 

\begin{lstlisting}
class StateEncoder(nn.Module):
    def __init__(self, state_dim, out_dim):
        super().__init__()
        # 3-layer MLP: Input -> 512 -> 512 -> Output
        self.net = MLP([state_dim, 512, 512, out_dim], 
                       layer_norm=True, output_activation=nn.GELU)

    def forward(self, state):
        return self.net(state)
\end{lstlisting}

\paragraph{Multi-scale Predictive Architecture}

\begin{lstlisting}
class MSPR(nn.Module):
    def __init__(self,
                 zs_dim: int,
                 action_dim: int,
                 state_encoder: nn.Module,
                 za_dim: int = 256,
                 zsa_dim: int = 512):
        super().__init__()
        self.zs = state_encoder()   # zs
        
        goal_dim = zs_dim      # zg uses the same encoder 
        
        # f_dyn: zsa -> next zs
        self.model = MLP(sizes=[zsa_dim, zs_dim], activation=nn.GELU, 
                        layer_norm=False)
        # f_rew: (zsa, phi) -> return
        self.rew_model = MLP(sizes=[zsa_dim + goal_dim, 1], 
                        activation=nn.GELU, layer_norm=False)

        # zsa
        self.za = MLP(sizes=[action_dim, za_dim], activation=nn.GELU,
                        output_activation=nn.GELU, layer_norm=False)
        self.zsa = MLP(sizes=[zs_dim + za_dim, 512, 512, zsa_dim], 
                        activation=nn.GELU, layer_norm=True)

        # f_g-act: (zs, zg) -> action
        self.high_inv_model = MLP(sizes=[zs_dim+goal_dim, action_dim], 
                                activation=nn.GELU, layer_norm=False)
        # f_g-dyn: (zs, zg) -> next zs
        self.high_dyn_model = MLP(sizes=[zs_dim+goal_dim, zs_dim], 
                                activation=nn.GELU, layer_norm=False)

    def forward(self, zs, action):
        za = self.za(action)
        return self.zsa(torch.cat([zs, za], -1))

    def model_all(self, zs, zg, action):
        zsa = self.forward(zs, action)

        pred_next_zs = self.model(zsa)
        pred_reward = self.rew_model(torch.cat([zsa, zg],-1))
        pred_action_high = self.high_inv_model(torch.cat([zs, zg],-1))
        pred_future_zs = self.high_dyn_model(torch.cat([zs, zg],-1))

        return pred_next_zs, pred_action_high, pred_reward, pred_future_zs
\end{lstlisting}
\paragraph{Value $Q$ Networks}

The value network ($Q$ or $V$). It takes the concatenated state-action representation $z_{sa}$ and goal representation $z_g$, and outputs the value of the state-action pair given the goal.

\begin{lstlisting}
class Value(nn.Module):
    def __init__(self, state_dim: int, 
                 hidden_dims: Sequence[int] = [512, 512, 512],
                 ):
        super().__init__()

        class ValueNetwork(nn.Module):
            def __init__(self, state_dim):
                super().__init__()
                self.mlp_module = MLP(sizes=[state_dim]+hidden_dims + [1], activation=nn.GELU, layer_norm=True)

            def forward(self, zsa, goal):
                return self.mlp_module(torch.concat([zsa, goal], -1))

        self.q1 = ValueNetwork(state_dim)
        self.q2 = ValueNetwork(state_dim)

    def forward(self, zsa, zg):
        return self.q1(zsa, goal), self.q2(zsa, goal)
\end{lstlisting}

\paragraph{Policy $\pi$ Network}

The policy network $\pi$ processes the state representation $z_s$ and goal representation $z_g$ the output is squashed via a Tanh activation.

\begin{lstlisting}
class Actor(nn.Module):
    def __init__(self, state_dim: int, action_dim: int,
                 hidden_dims: Sequence[int] = [512, 512, 512], tansquash=True):
        super().__init__()
        self.actor_net = MLP(sizes=[state_dim] + hidden_dims, activation=nn.GELU, output_activation=nn.GELU)
        self.mean_net = nn.Linear(512, action_dim)
        apply_variance_scaling_init(self.mean_net, scale=0.01)
        self.activ = nn.Tanh() if tansquash else nn.Identity()

    def forward(self, zs: torch.Tensor, goal: torch.Tensor):
        outputs = self.actor_net(torch.concat([zs, goal], -1))
        return self.activ(self.mean_net(outputs))
\end{lstlisting}

\subsection{Computational Cost}

All experiments were conducted on NVIDIA RTX 4090 GPUs / MIG partitions of H100s. We report approximate wall-clock training times per run for \ours{} across state-based and pixel-based settings on 4090 Machine. \ours{} requires $\sim 3.5$ hours for state-based tasks and $\sim 4.5$ hours for pixel-based tasks. The additional time stems from the multi-scale predictive modules trained over the horizon $H$. Notably, the increase in cost when moving from state to pixel inputs is modest ($+1$\,hr).

\section{Baseline Details}
\label{sec:baseline-details}
\subsection{GCRL Baselines.}

We evaluate against the standard offline GCRL algorithms provided in the OGBench codebase~\citep{park2024ogbench}: \textbf{GCBC}, \textbf{GCIQL}, \textbf{GCIVL}, \textbf{CRL}, \textbf{QRL}, and \textbf{HIQL}. We use the official implementations and hyperparameter configurations as reported in~\citet{park2024ogbench} without modification, except for the $n$-step targets of the TD-based methods (Appendix~\ref{app:nstep}).

\paragraph{Representation Learning Baselines.}
All Dual Goal Representation~\citep{dual_goal} baselines are taken from the official repository at commit \texttt{516c73be6cd470c8748fea6550b609ce0eedd30c}. The original authors provide a pixel-based implementation for \textbf{GCIVL} only; we implemented the pixel-based version of \textbf{Dual-CRL} ourselves. All extended implementations follow the same architectural and training conventions as the original Dual codebase to ensure a fair comparison. Although there is a discrepancy between the scores reported by OGBench \citep{park2024ogbench} for GCIVL and the original scores in Dual Goal Representation \citep{dual_goal}, we report the original scores from OGBench and re-run the Dual backbones for consistency in our comparisons.

\paragraph{$n$-step Targets}
\label{app:nstep}
To implement the $n$-step versions of HIQL and GCIVL, we follow the implementation details and official codebase of~\citet{park_long_horizon}.

\section{Additional Results}

\subsection{GCTD3 vs.\ \ours{}: Per-Environment Results}
\label{app:td3}
Tables~\ref{tab:td3} and~\ref{tab:td3_visual} report the per-environment numbers behind Figure~\ref{fig:td3_backbone} (Sec-\ref{sec:exp_attribution}). GCTD3 is the TD3+BC backbone of \ours{} with OGBench-style networks: the actor and critic each learn their own features from their own losses, without the shared predictive encoder. The agent, optimizer, $n$-step targets, and evaluation protocol are the same as for \ours{}. 

\begin{table}[ht]
\centering
\begin{adjustbox}{width=0.7\textwidth,keepaspectratio}
\begin{tabular}{l |cc |cc}
\toprule
 & \multicolumn{2}{c}{$\boldsymbol{{n=1}}$} & \multicolumn{2}{c}{$\boldsymbol{{n=3}}$} \\
\cmidrule(lr){2-3}\cmidrule(lr){4-5}
\textbf{Environment} & \textbf{TD3} & \textbf{OURS} & \textbf{TD3} & \textbf{OURS} \\
\midrule
\texttt{pointmaze-medium-navigate}
  & $23$ {\tiny $\pm 9$} & $\mathbf{50}$ {\tiny $\pm 1$} & $\mathbf{77}$ {\tiny $\pm 6$} & $54$ {\tiny $\pm 6$} \\
\texttt{pointmaze-large-navigate}
  & $0$ {\tiny $\pm 0$} & $\mathbf{30}$ {\tiny $\pm 2$} & $37$ {\tiny $\pm 2$} & $\mathbf{48}$ {\tiny $\pm 8$} \\
\texttt{antmaze-medium-navigate}
  & $64$ {\tiny $\pm 5$} & $\mathbf{98}$ {\tiny $\pm 1$} & $79$ {\tiny $\pm 3$} & $\mathbf{99}$ {\tiny $\pm 1$} \\
\texttt{antmaze-large-navigate}
  & $22$ {\tiny $\pm 2$} & $\mathbf{96}$ {\tiny $\pm 1$} & $25$ {\tiny $\pm 4$} & $\mathbf{98}$ {\tiny $\pm 2$} \\
\texttt{antmaze-giant-navigate}
  & $0$ {\tiny $\pm 0$} & $\mathbf{43}$ {\tiny $\pm 4$} & $0$ {\tiny $\pm 0$} & $\mathbf{68}$ {\tiny $\pm 7$} \\
\texttt{humanoidmaze-medium-navigate}
  & $23$ {\tiny $\pm 4$} & $\mathbf{35}$ {\tiny $\pm 3$} & $26$ {\tiny $\pm 3$} & $\mathbf{67}$ {\tiny $\pm 5$} \\
\texttt{humanoidmaze-large-navigate}
  & $1$ {\tiny $\pm 1$} & $\mathbf{11}$ {\tiny $\pm 3$} & $6$ {\tiny $\pm 3$} & $\mathbf{58}$ {\tiny $\pm 3$} \\
\texttt{cube-single-play}
  & $\mathbf{58}$ {\tiny $\pm 7$} & $46$ {\tiny $\pm 4$} & $\mathbf{63}$ {\tiny $\pm 4$} & $\mathbf{65}$ {\tiny $\pm 5$} \\
\texttt{cube-double-play}
  & $10$ {\tiny $\pm 4$} & $\mathbf{11}$ {\tiny $\pm 2$} & $11$ {\tiny $\pm 1$} & $\mathbf{13}$ {\tiny $\pm 3$} \\
\texttt{scene-play}
  & $27$ {\tiny $\pm 4$} & $\mathbf{30}$ {\tiny $\pm 4$} & $42$ {\tiny $\pm 5$} & $\mathbf{45}$ {\tiny $\pm 5$} \\
\texttt{puzzle-3x3-play}
  & $\mathbf{96}$ {\tiny $\pm 3$} & $81$ {\tiny $\pm 4$} & $\mathbf{96}$ {\tiny $\pm 2$} & $\mathbf{98}$ {\tiny $\pm 1$} \\
\texttt{puzzle-4x4-play}
  & $63$ {\tiny $\pm 7$} & $\mathbf{73}$ {\tiny $\pm 2$} & $88$ {\tiny $\pm 4$} & $\mathbf{93}$ {\tiny $\pm 5$} \\
\midrule
\textbf{Average}
  & $32$ {\tiny $\pm 5$} & $\mathbf{50}$ {\tiny $\pm 3$} & $46$ {\tiny $\pm 4$} & $\mathbf{67}$ {\tiny $\pm 5$} \\
\bottomrule
\end{tabular}
\end{adjustbox}

\vspace{3pt}
\caption{\label{tab:td3}\textbf{GCTD3 backbone vs.\ \ours{}, state-based environments}, at $n=1$ and $n=3$ target horizons. Mean $\pm$ std over 8 seeds.}
\vspace{-5pt}
\end{table}

\begin{table}[h]
\centering
\begin{adjustbox}{width=0.9\textwidth,keepaspectratio}
\begin{tabular}{l |cc |cc |cc}
\toprule
 & \multicolumn{2}{c}{$\boldsymbol{{n=1}}$, \textbf{w/o FS}} & \multicolumn{2}{c}{$\boldsymbol{{n=3}}$, \textbf{w/o FS}} & \multicolumn{2}{c}{$\boldsymbol{{n=3}}$, \textbf{FS}} \\
\cmidrule(lr){2-3}\cmidrule(lr){4-5}\cmidrule(lr){6-7}
\textbf{Environment} & \textbf{TD3} & \textbf{OURS} & \textbf{TD3} & \textbf{OURS} & \textbf{TD3} & \textbf{OURS} \\
\midrule
\vamn
  & $80$ {\tiny $\pm 1$} & $\mathbf{95}$ {\tiny $\pm 1$} & $\mathbf{84}$ {\tiny $\pm 6$} & $\mathbf{86}$ {\tiny $\pm 4$} & $85$ {\tiny $\pm 8$} & $\mathbf{90}$ {\tiny $\pm 3$} \\
\valn
  & $30$ {\tiny $\pm 9$} & $\mathbf{81}$ {\tiny $\pm 9$} & $41$ {\tiny $\pm 1$} & $\mathbf{78}$ {\tiny $\pm 5$} & $50$ {\tiny $\pm 4$} & $\mathbf{81}$ {\tiny $\pm 4$} \\
\vcsp
  & $30$ {\tiny $\pm 5$} & $\mathbf{42}$ {\tiny $\pm 5$} & $30$ {\tiny $\pm 9$} & $\mathbf{63}$ {\tiny $\pm 1$} & $45$ {\tiny $\pm 8$} & $\mathbf{82}$ {\tiny $\pm 5$} \\
\vsp
  & $\mathbf{18}$ {\tiny $\pm 5$} & $16$ {\tiny $\pm 3$} & $20$ {\tiny $\pm 3$} & $\mathbf{25}$ {\tiny $\pm 3$} & $30$ {\tiny $\pm 6$} & $\mathbf{56}$ {\tiny $\pm 5$} \\
\vptp
  & $\mathbf{20}$ {\tiny $\pm 7$} & $13$ {\tiny $\pm 2$} & $\mathbf{36}$ {\tiny $\pm 11$} & $21$ {\tiny $\pm 1$} & $\mathbf{40}$ {\tiny $\pm 7$} & $30$ {\tiny $\pm 4$} \\
\midrule
\textbf{Average}
  & $36$ {\tiny $\pm 6$} & $\mathbf{49}$ {\tiny $\pm 5$} & $42$ {\tiny $\pm 7$} & $\mathbf{55}$ {\tiny $\pm 3$} & $50$ {\tiny $\pm 6$} & $\mathbf{68}$ {\tiny $\pm 4$} \\
\bottomrule
\end{tabular}
\end{adjustbox}

\vspace{3pt}
\caption{\label{tab:td3_visual}\textbf{GCTD3 backbone vs.\ \ours{}, pixel-based environments}, with and without frame stacking (FS) at $n=3$, and without FS at $n=1$. Mean $\pm$ std over 4 seeds.}
\vspace{-5pt}
\end{table}

\subsection{Results at Target Horizon (n=1)}
\label{app:qh1_results}
Table~\ref{tab:qh1_results} reports all methods at the published $n=1$ configuration, with \ours{} also trained at $n=1$. Moving from $n=1$ to $n=3$ raises \ours{} from 50 to 67 (state) and 61 to 68 (pixel), which is why Table~\ref{tab:full_results} re-runs every TD-based baseline at $n=3$.

\begin{table*}[h]
\centering
\begin{adjustbox}{width=\textwidth,keepaspectratio}
\begin{tabular}{l *{5}{c}  *{3}{c} c}
\toprule
\textbf{Environment}
  & \textbf{GCBC} & \textbf{GCIVL} & \textbf{GCIQL} & \textbf{QRL} & \textbf{CRL}
  & $\textbf{\small Dual}_{\textbf{\tiny GCIVL}}$ & $\textbf{\small Dual}_{\textbf{\tiny CRL}}$
  & \textbf{HIQL}
  & \textbf{OURS} \\
\midrule
\rowcolor{sb_orange!20}\multicolumn{10}{c}{\textbf{ State-based environments}} \\
\texttt{pointmaze-medium-navigate}
  & $9$ {\tiny $\pm 6$} & $63$ {\tiny $\pm 6$} & $53$ {\tiny $\pm 8$} & $\mathbf{82}$ {\tiny $\pm 5$} & $29$ {\tiny $\pm 7$}
  & $76$ {\tiny $\pm 7$} & $33$ {\tiny $\pm 10$}
  & $\mathbf{79}$ {\tiny $\pm 5$}
  & $50$ {\tiny $\pm 1$} \\
\texttt{pointmaze-large-navigate}
  & $29$ {\tiny $\pm 6$} & $45$ {\tiny $\pm 5$} & $34$ {\tiny $\pm 3$} & $\mathbf{86}$ {\tiny $\pm 9$} & $39$ {\tiny $\pm 7$}
  & $46$ {\tiny $\pm 6$} & $39$ {\tiny $\pm 12$}
  & $58$ {\tiny $\pm 5$}
  & $30$ {\tiny $\pm 2$} \\
\texttt{antmaze-medium-navigate}
  & $29$ {\tiny $\pm 4$} & $72$ {\tiny $\pm 8$} & $71$ {\tiny $\pm 4$} & $88$ {\tiny $\pm 3$} & $\mathbf{95}$ {\tiny $\pm 1$}
  & $75$ {\tiny $\pm 4$} & $93$ {\tiny $\pm 3$}
  & $\mathbf{96}$ {\tiny $\pm 1$}
  & $\mathbf{98}$ {\tiny $\pm 1$} \\
\texttt{antmaze-large-navigate}
  & $24$ {\tiny $\pm 2$} & $16$ {\tiny $\pm 5$} & $34$ {\tiny $\pm 4$} & $75$ {\tiny $\pm 6$} & $83$ {\tiny $\pm 4$}
  & $28$ {\tiny $\pm 11$} & $87$ {\tiny $\pm 2$}
  & $91$ {\tiny $\pm 2$}
  & $\mathbf{96}$ {\tiny $\pm 1$} \\
\texttt{antmaze-giant-navigate}
  & $0$ {\tiny $\pm 0$} & $0$ {\tiny $\pm 0$} & $0$ {\tiny $\pm 0$} & $14$ {\tiny $\pm 3$} & $16$ {\tiny $\pm 3$}
  & $0$ {\tiny $\pm 0$} & $21$ {\tiny $\pm 4$}
  & $\mathbf{65}$ {\tiny $\pm 5$}
  & $43$ {\tiny $\pm 4$} \\
\texttt{humanoidmaze-medium-navigate}
  & $8$ {\tiny $\pm 2$} & $24$ {\tiny $\pm 2$} & $27$ {\tiny $\pm 2$} & $21$ {\tiny $\pm 8$} & $60$ {\tiny $\pm 4$}
  & $29$ {\tiny $\pm 3$} & $57$ {\tiny $\pm 4$}
  & $\mathbf{89}$ {\tiny $\pm 2$}
  & $35$ {\tiny $\pm 3$} \\
\texttt{humanoidmaze-large-navigate}
  & $1$ {\tiny $\pm 0$} & $2$ {\tiny $\pm 1$} & $2$ {\tiny $\pm 1$} & $5$ {\tiny $\pm 1$} & $24$ {\tiny $\pm 4$}
  & $3$ {\tiny $\pm 2$} & $18$ {\tiny $\pm 4$}
  & $\mathbf{49}$ {\tiny $\pm 4$}
  & $11$ {\tiny $\pm 3$} \\
\texttt{cube-single-play}
  & $6$ {\tiny $\pm 2$} & $53$ {\tiny $\pm 4$} & $68$ {\tiny $\pm 6$} & $5$ {\tiny $\pm 1$} & $19$ {\tiny $\pm 2$}
  & $\mathbf{89}$ {\tiny $\pm 3$} & $60$ {\tiny $\pm 1$}
  & $15$ {\tiny $\pm 3$}
  & $46$ {\tiny $\pm 4$} \\
\texttt{cube-double-play}
  & $1$ {\tiny $\pm 1$} & $36$ {\tiny $\pm 3$} & $40$ {\tiny $\pm 5$} & $1$ {\tiny $\pm 0$} & $10$ {\tiny $\pm 2$}
  & $\mathbf{60}$ {\tiny $\pm 4$} & $24$ {\tiny $\pm 5$}
  & $6$ {\tiny $\pm 2$}
  & $11$ {\tiny $\pm 2$} \\
\texttt{scene-play}
  & $5$ {\tiny $\pm 1$} & $42$ {\tiny $\pm 4$} & $51$ {\tiny $\pm 4$} & $5$ {\tiny $\pm 1$} & $19$ {\tiny $\pm 2$}
  & $\mathbf{72}$ {\tiny $\pm 6$} & $44$ {\tiny $\pm 5$}
  & $38$ {\tiny $\pm 3$}
  & $30$ {\tiny $\pm 4$} \\
\texttt{puzzle-3x3-play}
  & $2$ {\tiny $\pm 0$} & $6$ {\tiny $\pm 1$} & $\mathbf{95}$ {\tiny $\pm 1$} & $1$ {\tiny $\pm 0$} & $3$ {\tiny $\pm 1$}
  & $5$ {\tiny $\pm 1$} & $6$ {\tiny $\pm 1$}
  & $12$ {\tiny $\pm 2$}
  & $81$ {\tiny $\pm 4$} \\
\texttt{puzzle-4x4-play}
  & $0$ {\tiny $\pm 0$} & $13$ {\tiny $\pm 2$} & $26$ {\tiny $\pm 3$} & $0$ {\tiny $\pm 0$} & $0$ {\tiny $\pm 0$}
  & $23$ {\tiny $\pm 3$} & $2$ {\tiny $\pm 0$}
  & $7$ {\tiny $\pm 2$}
  & $\mathbf{73}$ {\tiny $\pm 2$} \\
\midrule
\textbf{Avg. State}
  & $10$ {\tiny $\pm 1$} & $31$ {\tiny $\pm 1$} & $42$ {\tiny $\pm 1$} & $32$ {\tiny $\pm 1$} & $33$ {\tiny $\pm 1$}
  & $42$ {\tiny $\pm 1$} & $40$ {\tiny $\pm 2$}
  & $\mathbf{50}$ {\tiny $\pm 1$}
  & $\mathbf{50}$ {\tiny $\pm 1$} \\
\midrule
\rowcolor{sb_orange!20}\multicolumn{10}{c}{\textbf{Pixel-based environments}} \\
\vamn
  & $11$ {\tiny $\pm 2$} & $22$ {\tiny $\pm 2$} & $11$ {\tiny $\pm 1$} & $0$ {\tiny $\pm 0$} & $\mathbf{95}$ {\tiny $\pm 1$}
  & $78$ {\tiny $\pm 4$} & $85$ {\tiny $\pm 3$}
  & $\mathbf{93}$ {\tiny $\pm 4$}
  & $\mathbf{92}$ {\tiny $\pm 3$} \\
\valn
  & $4$ {\tiny $\pm 0$} & $5$ {\tiny $\pm 1$} & $4$ {\tiny $\pm 1$} & $0$ {\tiny $\pm 0$} & $\mathbf{84}$ {\tiny $\pm 1$}
  & $40$ {\tiny $\pm 4$} & $43$ {\tiny $\pm 3$}
  & $53$ {\tiny $\pm 9$}
  & $77$ {\tiny $\pm 5$} \\
\vcsp
  & $5$ {\tiny $\pm 1$} & $60$ {\tiny $\pm 4$} & $30$ {\tiny $\pm 5$} & $41$ {\tiny $\pm 15$} & $31$ {\tiny $\pm 15$}
  & $58$ {\tiny $\pm 5$} & $14$ {\tiny $\pm 8$}
  & $\mathbf{89}$ {\tiny $\pm 0$}
  & $82$ {\tiny $\pm 2$} \\
\vsp
  & $12$ {\tiny $\pm 2$} & $25$ {\tiny $\pm 2$} & $12$ {\tiny $\pm 2$} & $10$ {\tiny $\pm 1$} & $11$ {\tiny $\pm 2$}
  & $26$ {\tiny $\pm 5$} & $15$ {\tiny $\pm 1$}
  & $\mathbf{49}$ {\tiny $\pm 4$}
  & $36$ {\tiny $\pm 6$} \\
\vptp
  & $0$ {\tiny $\pm 0$} & $21$ {\tiny $\pm 2$} & $1$ {\tiny $\pm 2$} & $1$ {\tiny $\pm 1$} & $0$ {\tiny $\pm 0$}
  & $0$ {\tiny $\pm 0$} & $0$ {\tiny $\pm 0$}
  & $\mathbf{73}$ {\tiny $\pm 8$}
  & $20$ {\tiny $\pm 1$} \\
\midrule
\textbf{Avg. Pixel}
  & $6$ {\tiny $\pm 1$} & $27$ {\tiny $\pm 1$} & $12$ {\tiny $\pm 1$} & $10$ {\tiny $\pm 3$} & $44$ {\tiny $\pm 3$}
  & $40$ {\tiny $\pm 2$} & $31$ {\tiny $\pm 2$}
  & $\mathbf{71}$ {\tiny $\pm 3$}
  & $61$ {\tiny $\pm 2$} \\
\bottomrule
\end{tabular}
\end{adjustbox}

\vspace{3pt}
\caption{\label{tab:qh1_results}\textbf{OGBench results at $n=1$.} Baselines use their published configurations; \ours{} uses $n=1$ targets (with frame stacking on pixel). Same layout and bold rule as Table~\ref{tab:full_results}.}
\end{table*}

\subsection{Results on Stitch Datasets}
\label{app:stitch}
Table~\ref{tab:stitch_results} reports results on the OGBench \texttt{stitch} datasets, whose trajectories are short fragments that individually do not reach the evaluation goals, so success requires composing behavior across trajectory boundaries. All methods use the same configurations as in Table~\ref{tab:full_results}. \ours{} outperforms both Dual goal-representation methods on every environment, by 22 points on \texttt{antmaze-medium-stitch} (78 vs.\ 56) and 46 on \texttt{visual-antmaze-medium-stitch} (83 vs.\ 37), and has the highest average among non-hierarchical methods in both settings (61 on state, 49 on pixel). HIQL is the strongest method on all five environments (87 and 57 on average), consistent with the advantage of subgoal decomposition when goals lie beyond the span of individual trajectories. Among the remaining baselines, QRL leads on \texttt{pointmaze-medium-stitch} (80 vs.\ 58), where its quasimetric objective matches the purely geometric maze, and GCBC reaches 24 on \texttt{visual-antmaze-large-stitch}, where all other methods except HIQL remain at or below 15.

\begin{table*}[h]
\centering
\begin{adjustbox}{width=\textwidth,keepaspectratio}
\begin{tabular}{l *{5}{c}  *{3}{c} c}
\toprule
\textbf{Environment}
  & \textbf{GCBC} & \textbf{GCIVL} & \textbf{GCIQL} & \textbf{QRL} & \textbf{CRL}
  & $\textbf{\small Dual}_{\textbf{\tiny GCIVL}}$ & $\textbf{\small Dual}_{\textbf{\tiny CRL}}$
  & \textbf{HIQL}
  & \textbf{OURS} \\
\midrule
\rowcolor{sb_orange!20}\multicolumn{10}{c}{\textbf{ State-based environments}} \\
\texttt{antmaze-medium-stitch}
  & $45$ {\tiny $\pm 11$} & $44$ {\tiny $\pm 5$} & $34$ {\tiny $\pm 3$} & $59$ {\tiny $\pm 7$} & $53$ {\tiny $\pm 2$}
  & $40$ {\tiny $\pm 3$} & $56$ {\tiny $\pm 4$}
  & $\mathbf{93}$ {\tiny $\pm 4$}
  & $78$ {\tiny $\pm 4$} \\
\texttt{humanoidmaze-medium-stitch}
  & $29$ {\tiny $\pm 5$} & $14$ {\tiny $\pm 2$} & $35$ {\tiny $\pm 6$} & $18$ {\tiny $\pm 2$} & $36$ {\tiny $\pm 6$}
  & $17$ {\tiny $\pm 2$} & $44$ {\tiny $\pm 3$}
  & $\mathbf{89}$ {\tiny $\pm 5$}
  & $46$ {\tiny $\pm 3$} \\
\texttt{pointmaze-medium-stitch}
  & $23$ {\tiny $\pm 18$} & $75$ {\tiny $\pm 5$} & $27$ {\tiny $\pm 5$} & $\mathbf{80}$ {\tiny $\pm 12$} & $0$ {\tiny $\pm 1$}
  & $51$ {\tiny $\pm 2$} & $2$ {\tiny $\pm 5$}
  & $\mathbf{78}$ {\tiny $\pm 3$}
  & $58$ {\tiny $\pm 3$} \\
\midrule
\textbf{Avg. State}
  & $32$ {\tiny $\pm 7$} & $44$ {\tiny $\pm 2$} & $32$ {\tiny $\pm 3$} & $52$ {\tiny $\pm 5$} & $30$ {\tiny $\pm 2$}
  & $36$ {\tiny $\pm 1$} & $34$ {\tiny $\pm 2$}
  & $\mathbf{87}$ {\tiny $\pm 2$}
  & $61$ {\tiny $\pm 2$} \\
\midrule
\rowcolor{sb_orange!20}\multicolumn{10}{c}{\textbf{Pixel-based environments}} \\
\vams
  & $64$ {\tiny $\pm 4$} & $16$ {\tiny $\pm 2$} & $12$ {\tiny $\pm 2$} & $0$ {\tiny $\pm 0$} & $69$ {\tiny $\pm 2$}
  & $26$ {\tiny $\pm 2$} & $37$ {\tiny $\pm 3$}
  & $\mathbf{86}$ {\tiny $\pm 2$}
  & $\mathbf{83}$ {\tiny $\pm 3$} \\
\vals
  & $24$ {\tiny $\pm 3$} & $1$ {\tiny $\pm 1$} & $0$ {\tiny $\pm 0$} & $1$ {\tiny $\pm 1$} & $11$ {\tiny $\pm 3$}
  & $1$ {\tiny $\pm 1$} & $10$ {\tiny $\pm 1$}
  & $\mathbf{28}$ {\tiny $\pm 2$}
  & $15$ {\tiny $\pm 3$} \\
\midrule
\textbf{Avg. Pixel}
  & $44$ {\tiny $\pm 3$} & $9$ {\tiny $\pm 1$} & $6$ {\tiny $\pm 1$} & $1$ {\tiny $\pm 1$} & $40$ {\tiny $\pm 2$}
  & $14$ {\tiny $\pm 1$} & $24$ {\tiny $\pm 2$}
  & $\mathbf{57}$ {\tiny $\pm 1$}
  & $49$ {\tiny $\pm 2$} \\
\bottomrule
\end{tabular}
\end{adjustbox}

\vspace{3pt}
\caption{\label{tab:stitch_results}\textbf{OGBench results on \texttt{stitch} datasets.} Same layout and bold rule as Table~\ref{tab:full_results}; TD-based methods use $n=3$.}
\end{table*}

\clearpage
\section{Analysis}
\label{sec:failure-case}
\begin{figure}[h]
    \centering
    \includegraphics[width=0.87\linewidth]{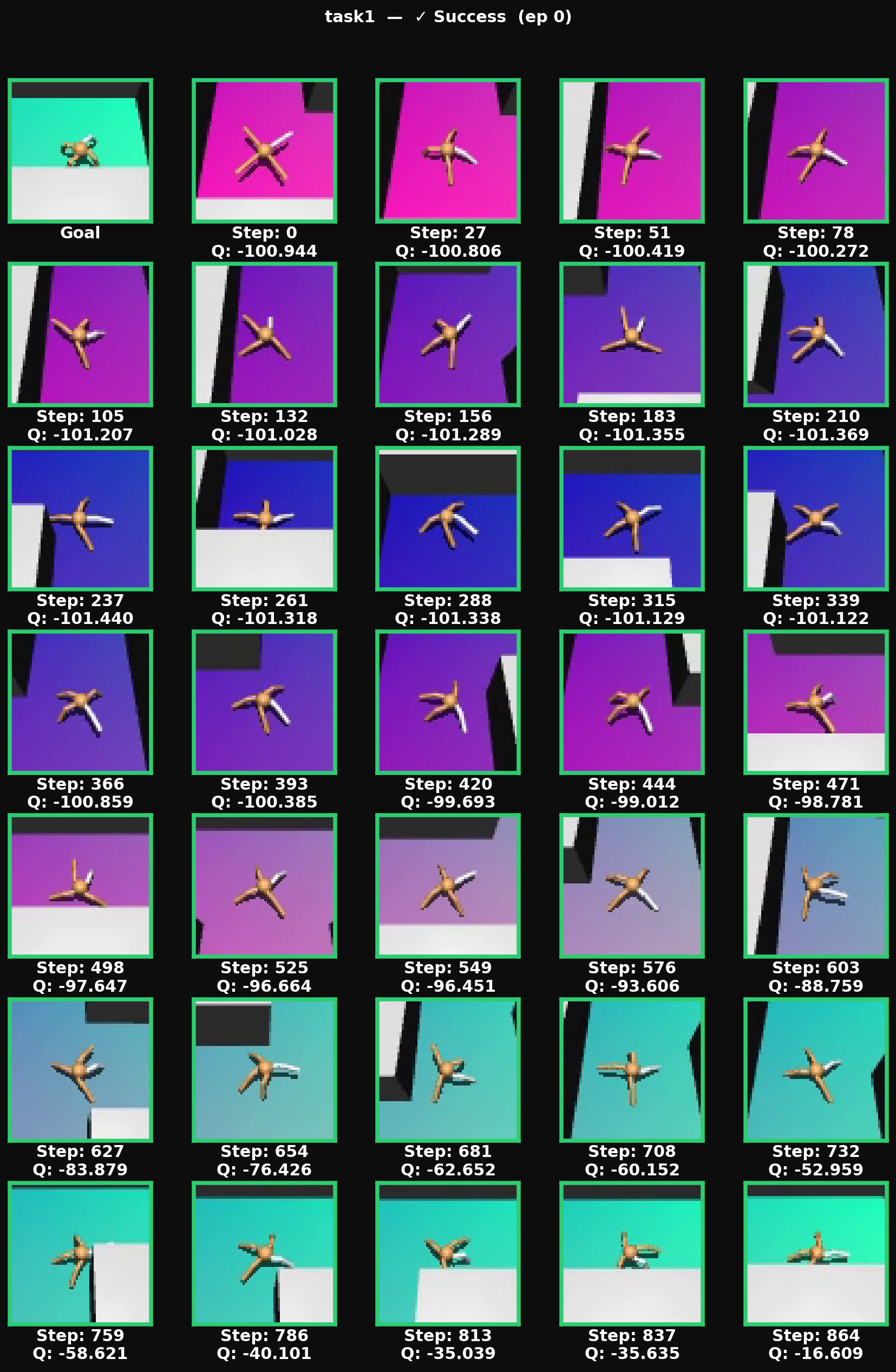}
    \caption{\textit{Successful Trajectory}: Goal in top-left. We show the progress, along with the Q values estimated by the Critic for \valn. }
    \label{fig:valn_success}
\end{figure}
\begin{figure}[ht]
    \centering
    \includegraphics[width=0.99\linewidth]{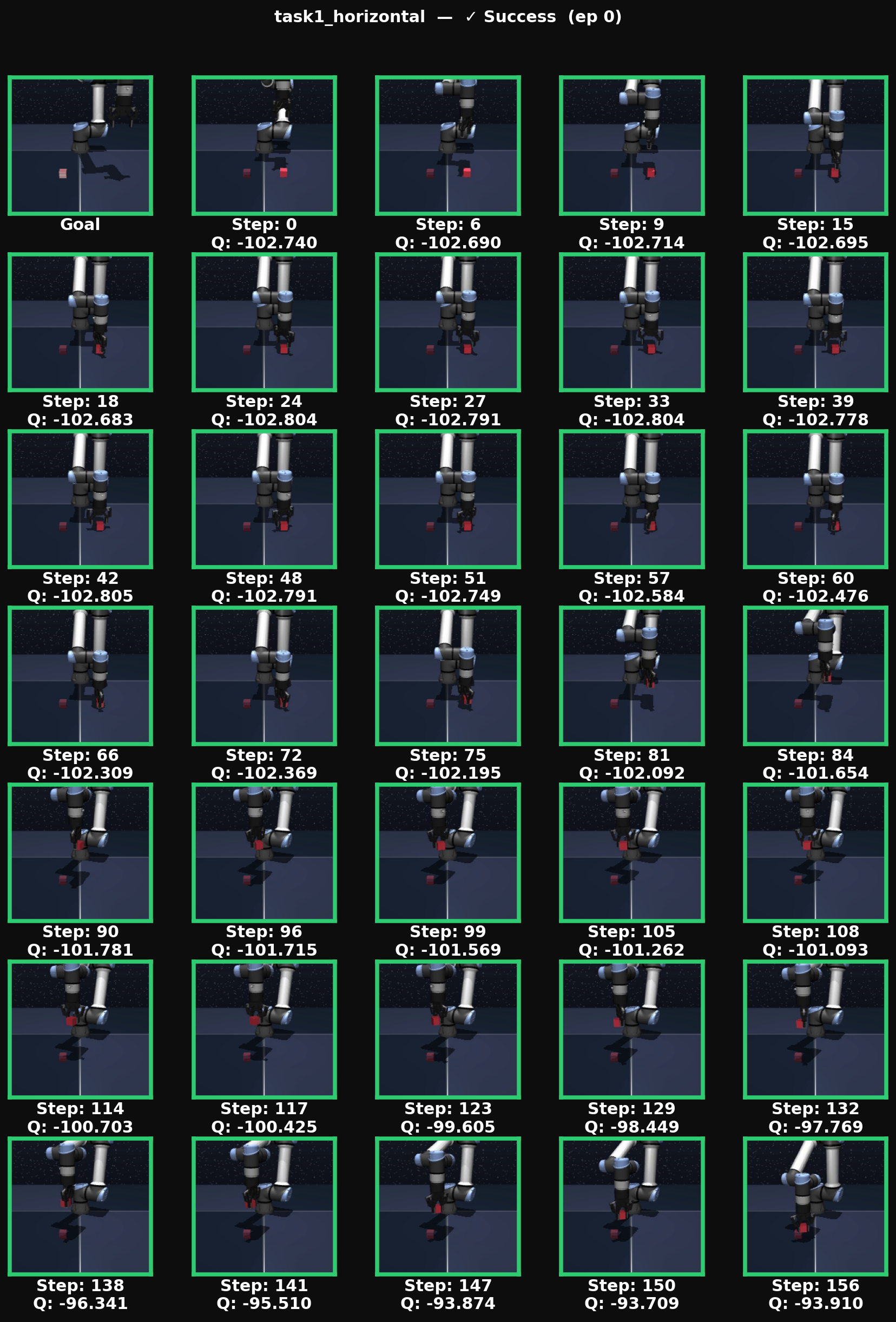}
    \caption{\textit{Successful case} (\texttt{Cube-Single-play}): Although the trajectory successfully reaches the goal, the Q-values do not reflect this, because of the behavior cloning term in the actor objective function}
    \label{fig:puzzle_success}
\end{figure}

\begin{figure}[ht]
    \centering
    \includegraphics[width=0.99\linewidth]{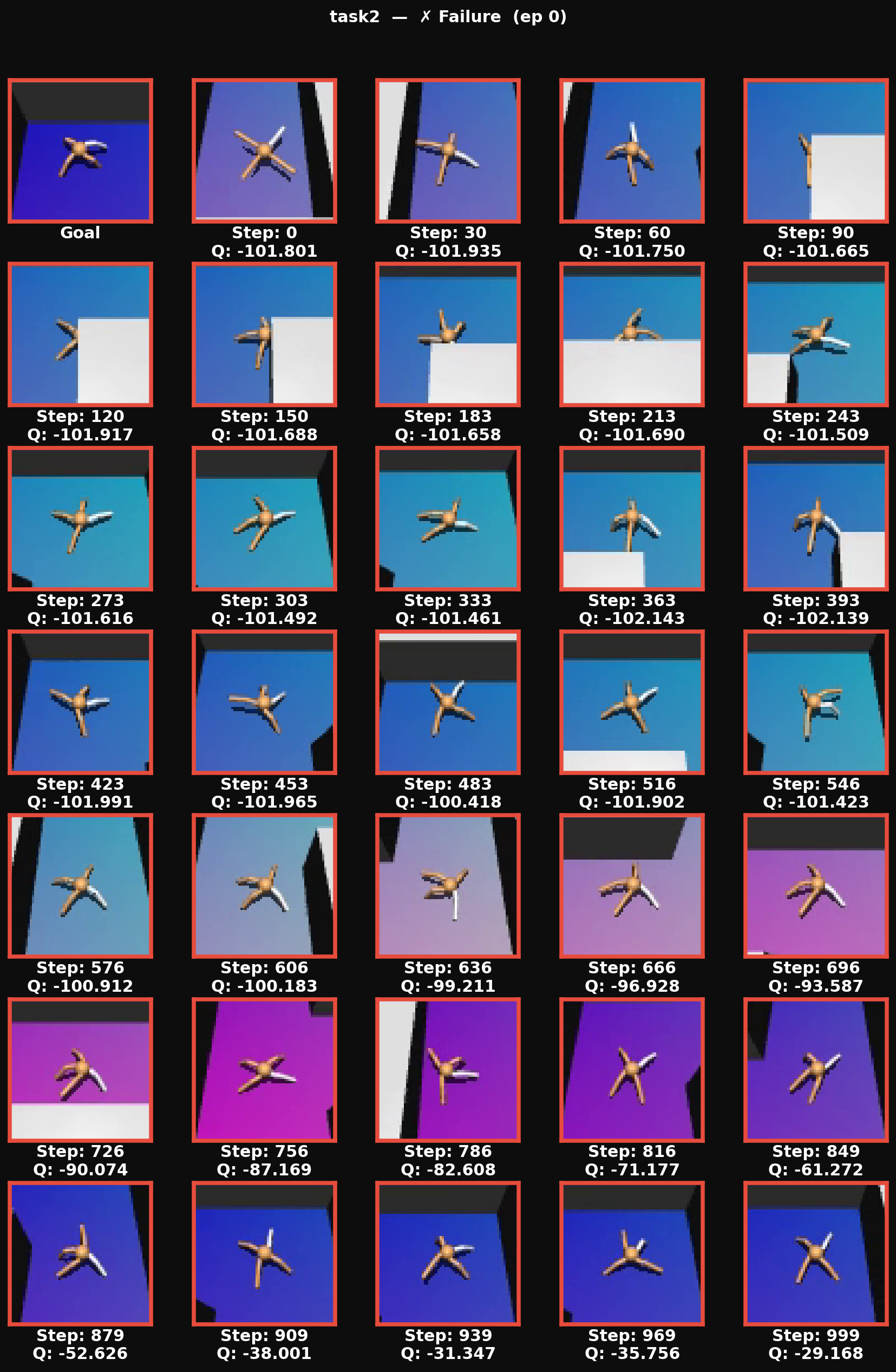}
    \caption{\textit{Failure case} : Goal in top-left. We show the progress, along with the Q values estimated by the Critic for \valn. Here the agent reaches the state which is similar to the goal but not exactly the goal  }
    \label{fig:valn_fail}
\end{figure}

\begin{figure}[ht]
    \centering
    \includegraphics[width=0.99\linewidth]{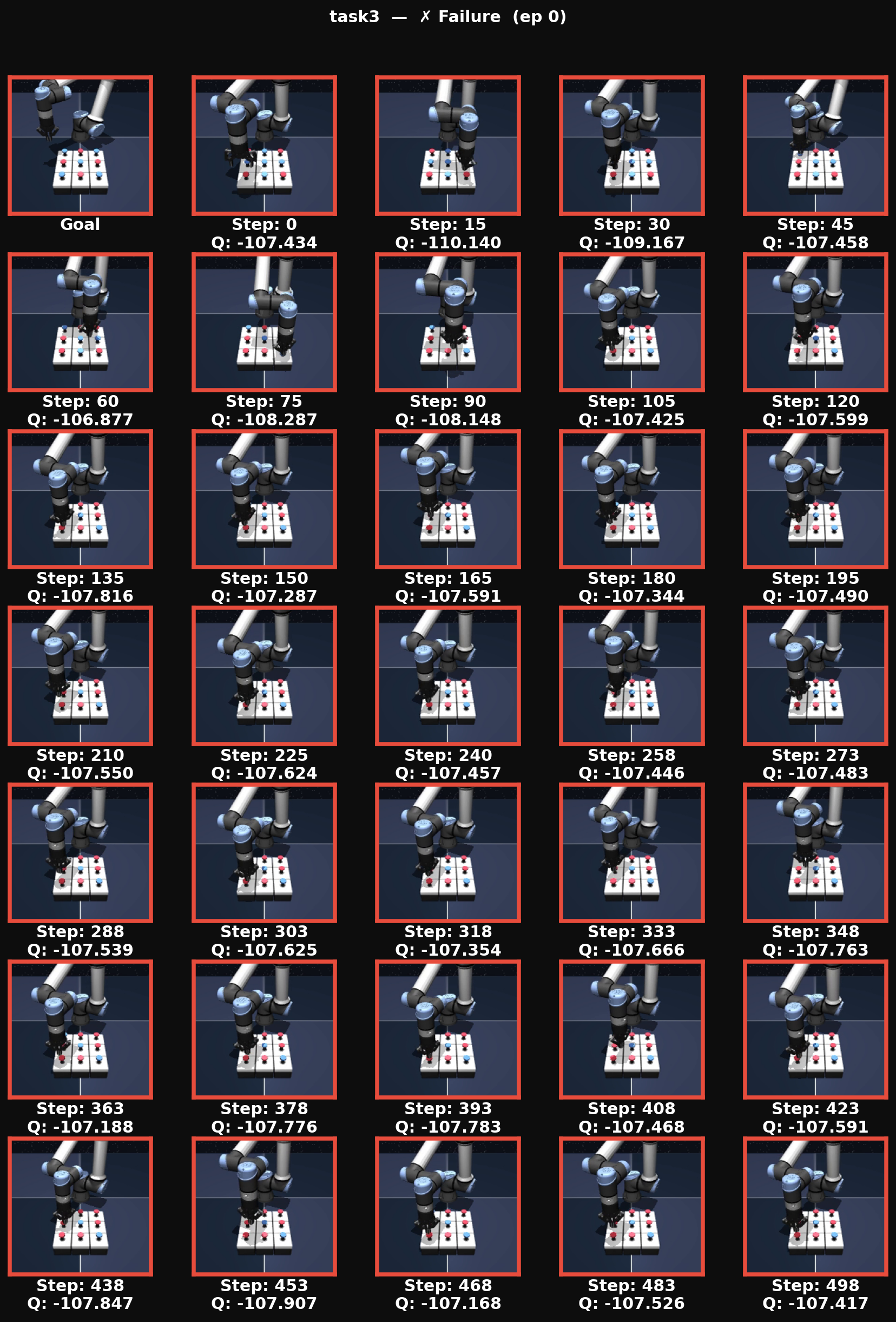}
    \caption{\textit{Failure case} (\texttt{puzzle-3x3-play}): Here from step ${\sim}200$ onwards, the agent enters a repetitive action loop without making meaningful progress, suggesting the policy has converged to a local behavioral mode that does not correspond to any goal-directed transition in the learned representation}
    \label{fig:puzzle_fail}
\end{figure}

\clearpage
\subsection*{AI use statement}

For this work, Generative AI tools were used to improve the writing, verifying the implementation, cross checking the math proofs. Generative AI tools were not used for literature retrieval and discovery, research ideation or execution, synthetic dataset generation, or proving mathematical claims.

\subsection*{Ethics statement}

There is no ethics concern for this work \ours{}. 

\subsection*{Reproducibility statement}

For full transparency and reproducibility, our implementation relies on publicly available codebases and benchmark suites. Detailed descriptions of key hyperparameters and Code Listings are provided in the Appendix to allow straightforward replication of our findings. 

\end{document}